\pdfoutput=1

\documentclass[11pt]{article}

\PassOptionsToPackage{hyphens,spaces,obeyspaces}{url}
\usepackage[final]{acl}

\usepackage{cleveref}

\usepackage{times}
\usepackage{latexsym}

\usepackage[T1]{fontenc}

\usepackage[utf8]{inputenc}

\usepackage{microtype}
\usepackage{graphicx}
\usepackage{soul}
\usepackage{subcaption}
\usepackage{inconsolata}
\usepackage{enumitem}
\title{When Debiasing Backfires: Counterintuitive Side Effects of Preprocessing-Based Stereotype Mitigation}

\author{Yahan Zheng \\
  Dartmouth College \\
  \texttt{yahan.zheng.gr@dartmouth.edu} \\\And
  John J. Guerrerio \\
  Dartmouth College \\
  \texttt{john.j.guerrerio.26@dartmouth.edu} \\\AND
  Soroush Vosoughi \\
  Dartmouth College \\
  \texttt{soroush.vosoughi@dartmouth.edu} \\\And
  Weicheng Ma  \\
  Oakland University \\
  \texttt{weichengma@oakland.edu} }
 
\begin{document}
\maketitle

\begin{abstract}
Preprocessing-based methods for stereotype mitigation, such as pre-/post-training on debiased corpora, are widely used in NLP. While these approaches reduce measurable stereotypes for targeted groups, we find they often induce unintended shifts—\emph{side effects}, where stereotyping or counter-stereotyping can \emph{increase} relative to neutral baselines for other demographics, including across unrelated demographic categories. We demonstrate these side effects across two model families (encoder-only and decoder-only), multiple preprocessing strategies (removing stereotypical sentences, removing group mentions, and swapping group references), and both pre- and post-training at different data scales on Wikipedia. Standard benchmarks frequently miss these shifts. Using attention-rollout analysis, we observe that such side effects are not accompanied by large changes in attention flow, complicating mechanistic explanations. We discuss implications for evaluation, provide actionable diagnostics, and argue for side-effect-aware, transparent mitigation practices.
\end{abstract}

\section{Introduction}
Pre-trained language models (PLMs) encode and propagate social stereotypes \citep{ManComputerProgrammer,SemanticsDerivedAutomatically,GenderBiasContextualized}, raising concerns about safe and responsible deployment. Among mitigation strategies, \emph{preprocessing}-based methods seek to modify training data, for instance, by removing or altering stereotypical content, because they are simple to apply and impose no inference-time cost \citep{StereotypeSurvey}. Intuitively, such interventions should hypothetically prevent models from learning unwanted associations by isolating them from stereotypical patterns.
For consistency, we use the term \emph{stereotype} to refer to any unwanted social bias encoded by language models from their training data, and \emph{mitigation} or \emph{debiasing} to refer to interventions aimed at reducing these encoded stereotypes.

Despite the intuitive appeal and partial effectiveness of preprocessing-based methods, to date, no PLM has achieved complete freedom from stereotypes, raising questions about their true efficacy.  More importantly, it is unclear whether data-level mitigation \emph{eliminates} harmful associations or merely \emph{redistributes} them. 

We revisit this question empirically by curating debiased Wikipedia corpora targeting six demographic groups spanning three categories (gender, race, religion). We use these corpora for both pre- and post-training of two PLMs (encoder-only TinyBERT \citep{tinybert-orig} and decoder-only GPT-2 \citep{gpt-2}) and then measure changes in stereotype expression toward \emph{all} groups. We report three findings:

\begin{enumerate}[leftmargin=*, noitemsep, topsep=0pt, parsep=0pt, partopsep=0pt]
\item \textbf{Unintended shifts within and across categories.} While stereotyping toward the group \emph{targeted} for debiasing decreases, we frequently observe increased stereotyping or counter-stereotyping for \emph{other} groups (including across bias categories). We define this phenomenon as a \textit{side effect}: a case where a mitigation method decreases stereotype measures for the target group but simultaneously induces undesired changes for one or more non-target groups. Notably, these shifts are often asymmetric and hard to anticipate. We quantify trends on StereoSet and CrowS-Pairs \citep{nadeem-etal-2021-stereoset,crows-orig}. These patterns cannot be explained solely by changes in the distribution of stereotypical/anti-stereotypical training examples.
\item \textbf{Robustness across settings.} Side effects appear under three preprocessing strategies (removing stereotypical sentences, removing group mentions, swapping references), during both pre- and post-training, \emph{and across training data scales}, for encoder-only (TinyBERT) and decoder-only (GPT-2) models. A larger decoder-only model exhibits the same qualitative phenomenon on a single post-training slice.
\item \textbf{Mechanism remains elusive.} Attention-rollout analysis \citep{attention-rollout} shows small attribution shifts even when stereotype scores move, suggesting that attention routing alone does not explain the effect and motivating distributional and causal follow-ups.
\end{enumerate}

Overall, our results highlight a reliability gap for preprocessing-based mitigation: interventions that help the target group can unpredictably harm others. These unintended consequences persist across diverse experimental conditions, including multiple preprocessing strategies, reduced training data scales, and different PLM architectures. In addition, standard evaluation benchmarks frequently fail to surface these side effects, and they are largely undetectable by inspecting models' internal attention patterns. Taken together, this \emph{calls into question the reliability and safety of data-level debiasing when used in isolation}. We recommend reporting side-effect–aware diagnostics (see Sections~\ref{sct:exp-positive}--\ref{sct:side-effects}) and emphasize the urgent need for \emph{more robust, interpretable, and controllable} mitigation methods capable of reducing social biases in language models. Our code for this paper has been publicly released at \url{https://github.com/InDaCS-Lab/Stereotype-Mitigation-Side-Effects}.

\section{Background}
Stereotype encoding and mitigation have been studied extensively in computational linguistics. Early work such as \citet{ManComputerProgrammer} demonstrated gender bias in Word2Vec embeddings trained on the Google News corpus. Shortly thereafter, \citet{SemanticsDerivedAutomatically} showed that GloVe embeddings likewise inherit human-like biases. Similar studies demonstrated these biases extended to contextualized models: \citet{GenderBiasContextualized} quantified and mitigated gender bias in ELMo's contextual word vectors. 
As larger, less interpretable Transformer-based models like BERT, GPT-2, and RoBERTa became dominant, the community shifted toward using specialized stereotype evaluation datasets, such as StereoSet~\cite{nadeem-etal-2021-stereoset}, WinoGender~\cite{winogender}, and CrowS-Pairs~\cite{crows-orig}, to benchmark bias in commonly-used language models. For instance, \citet{nadeem-etal-2021-stereoset} empirically demonstrated BERT and GPT-2 have stereotypical tendencies, a finding validated by \citet{crows-orig} for BERT. With the rise of Large Language Models (LLMs), prompt-based benchmarks such as \citet{ROBBIERobustBias}, \citet{BoldDatasetMetrics}, and \citet{akyurek-etal-2022-measuring} have been employed to examine biases in modern models.

The phenomenon of encoded stereotypes has prompted the creation of targeted mitigation strategies. As defined by \citet{StereotypeSurvey}, these mitigation approaches can be divided into 4 broad categories.

\textbf{Pre-Processing Mitigation} involves changes to the training data to prevent the model from learning stereotypes. One common form of preprocessing-based stereotype mitigation is data augmentation. \citet{GenderBiasNeural} first formalized this approach to mitigate gender bias by creating pairs of semantically invariant sentences with flipped gendered words (e.g., "he" to "she"). \citet{gendertuning} extended this approach by masking gendered words and using a language model to predict a replacement.
Another form of preprocessing-based stereotype mitigation approach is dataset filtering, which focuses on identifying examples to either emphasize or exclude. \citet{DemographicAwareLanguageModel} and \citet{LookingHandsomeCarpenter} identify underrepresented or low-bias examples to focus on during post-training. \citet{ExploringLimitsTransfer} uses a word list to filter out biased examples, an approach refined by \citet{MitigatingHarmLanguage} and \citet{FairInfinitesimalJackknife} with more advanced filtering techniques. \citet{ProxyClean} identifies demographic identifying words and removes them prior to post-training. Notably, data augmentation and data filtering can be combined; \citet{HealthyDiet} generates counterfactual examples for data that contribute the most to fairness and filters out other stereotypical examples.

\textbf{In-Training Mitigation} incorporates changes to the training procedure or additional post-training steps. For instance, \citet{autodebias}, \citet{AdeptDebiasingPrompt}, and \citet{DebiasingPretrainedText} introduce new loss functions to mitigate the biases the model potentially learns. \citet{efficientfinetuning} fine-tune a very small subset of model parameters on the WinoBias and CrowS-Pairs datasets. \citet{TrainingLanguageModels} employs a reinforcement learning-based post-training approach with human feedback to better align LLMs with human values.

\textbf{Intra-Processing Mitigation} entails modifications to the model's inference behavior. Works such as \citet{DecipheringStereotypesPreTrained} and  \citet{UnibiasUnveilingMitigating} identify biased model components (e.g., attention heads) and disable them at inference time. \citet{TowardsResourceEfficient} uses smaller stereotypical and anti-stereotypical expert models to re-balance next token probabilities toward anti-stereotypical tokens and away from stereotypical ones. Similarly, \citet{BOLTFastEnergy} learns small, tunable bias vectors at inference time to shift the model's logits away from toxic tokens. Finally, \citet{firsttheworst} employs beam search to find more diverse model outputs at inference time.

\textbf{Post-Processing Mitigation} encompasses modifications to the model's output text generation. \citet{TextStyleTransfer} frames debiasing as a style transfer problem, and uses LIME to identify biased keywords to be replaced via style transfer to a neutral domain. \citet{QueerPeopleAre} employ Shapley values to identify biased words in model output and re-prompt the LLM to rephrase the given sentence without those words.

\textbf{Preprocessing-based} methods offer several important advantages over other stereotype mitigation approaches. As described in
\citet{StereotypeSurvey}, these approaches only modify the input of the model during training. This allows debiasing of models without introducing additional constraints to the training process or requiring additional compute at inference time. However, they also have significant limitations. Many data augmentation techniques swap terms using word lists, which can be incomplete and change the semantic meaning of a sentence. \citet{StereotypeSurvey} argues this limitation is especially salient for words describing social groups. Assuming the interchangeability of social groups ignores the fact that stereotypes are nuanced and specific to each group. Similarly, removing or replacing identity words does not eliminate the harm within a stereotypical statement, but only redirects it toward a potentially irrelevant group. 

Similar limitations apply to data filtering techniques. Incomplete and misrepresentative word lists can lead to the removal of minority voices while leaving behind harmful documents. Such techniques can also introduce distributional imbalances into the training data, exacerbating bias. 

Our work presents a comprehensive investigation into the limitations of these methods, while also characterizing the side effects they introduce.

\section{Experimental Settings}
We investigate the encoded stereotypes of TinyBERT (4-layer, 14M parameters) and GPT-2 (124M parameters) to examine potential side effects of preprocessing-based stereotype mitigation methods. TinyBERT is an encoder-only transformer trained with a masked language modeling (MLM) objective, whereas GPT-2 is a decoder-only transformer trained with a causal language modeling (CLM) objective. Using two architecturally distinct yet compact PLMs provides diversity in the models we study (strengthening the generalizability of our findings) while keeping multiple rounds of pre-/post-training and evaluation computationally feasible.

All pre-/post-training uses the June~1,~2023 English Wikipedia snapshot, tokenized and filtered to prose articles. For TinyBERT, this domain matches the original training data used for the released model. For evaluation, we employ the intra-sentence portion of StereoSet and CrowS-Pairs. In addition to reporting overall scores, we report group-specific stereotype metrics for gender- (female and male), race- (Black and Caucasian), and religion- (Muslim and Christian) based demographic groups. This allows us to capture more fine-grained effects of stereotype mitigation.

We consider three commonly used data-level interventions. Note, we conduct each intervention solely for \emph{one} of the six demographic groups or three demographic categories we consider (referred to as the \emph{target group} or \emph{target category} respectively). After completing an intervention, we measure stereotyping toward all demographic groups.
\begin{enumerate}[leftmargin=*, noitemsep, topsep=0pt, parsep=0pt, partopsep=0pt]
\item \textbf{Debias-A-Group (DG):} remove sentences flagged as \emph{stereotypical} toward the target group using a pretrained sentence-level stereotype detector.\footnote{\url{https://huggingface.co/wu981526092/Sentence-Level-Stereotype-Detector}} On CrowS-Pairs, this detector attains F1=0.98 (precision=0.98; recall=0.98). 
\item \textbf{Remove-A-Group (RG):} remove \emph{all} sentences mentioning the target group, regardless of sentiment or potential stereotypes.
\item \textbf{Swap-References (SR):} replace all identity mentions within the target category. For gender stereotypes, we apply direct antonym swaps (e.g., ``female'' $\leftrightarrow$ ``male''). For racial and religious stereotypes, where clear antonyms do not exist, we define group mappings and generate replacements using a constrained LLM prompt that enforces grammaticality, number, and syntactic role preservation. Three NLP experts annotated these samples, confirming that 96 out of 100 instances were of satisfactory quality (receiving approval from at least two annotators).
\end{enumerate}

We run the full grid of the three strategies $\times$ two training stages (pre-/post-) $\times$ two data scales (100\% and 5\%), yielding 12 settings. This setup enables us to compare the effects of data scale and training stage on the expression of stereotypes.

Models are implemented with HuggingFace.\footnote{\url{https://huggingface.co}} All runs use a single RTX 2080 Ti GPU, and the random seed is set to 42.

\paragraph{Evaluation protocol.} We report overall StereoSet stereotype scores (SS; 50 is neutral; farther from 50 indicates stronger stereotyping/anti-stereotyping, which is undesirable), language modeling score (LMS), and iCAT (which balances SS and LMS), alongside demographic group-level SS. For CrowS-Pairs we report per-demographic category scores following the original work. Because absolute SS is known to be sensitive to instance composition, we emphasize \emph{directional changes} relative to the corresponding base model in matched conditions. Where applicable, we additionally analyze attribution via attention-rollout \citep{attention-rollout}.

\section{Data Debiasing Reduces Stereotypes} \label{sct:exp-positive}
\begin{figure*}[!h]
    \centering
    \begin{subfigure}[b]{.32\linewidth}
        \includegraphics[width=\linewidth]{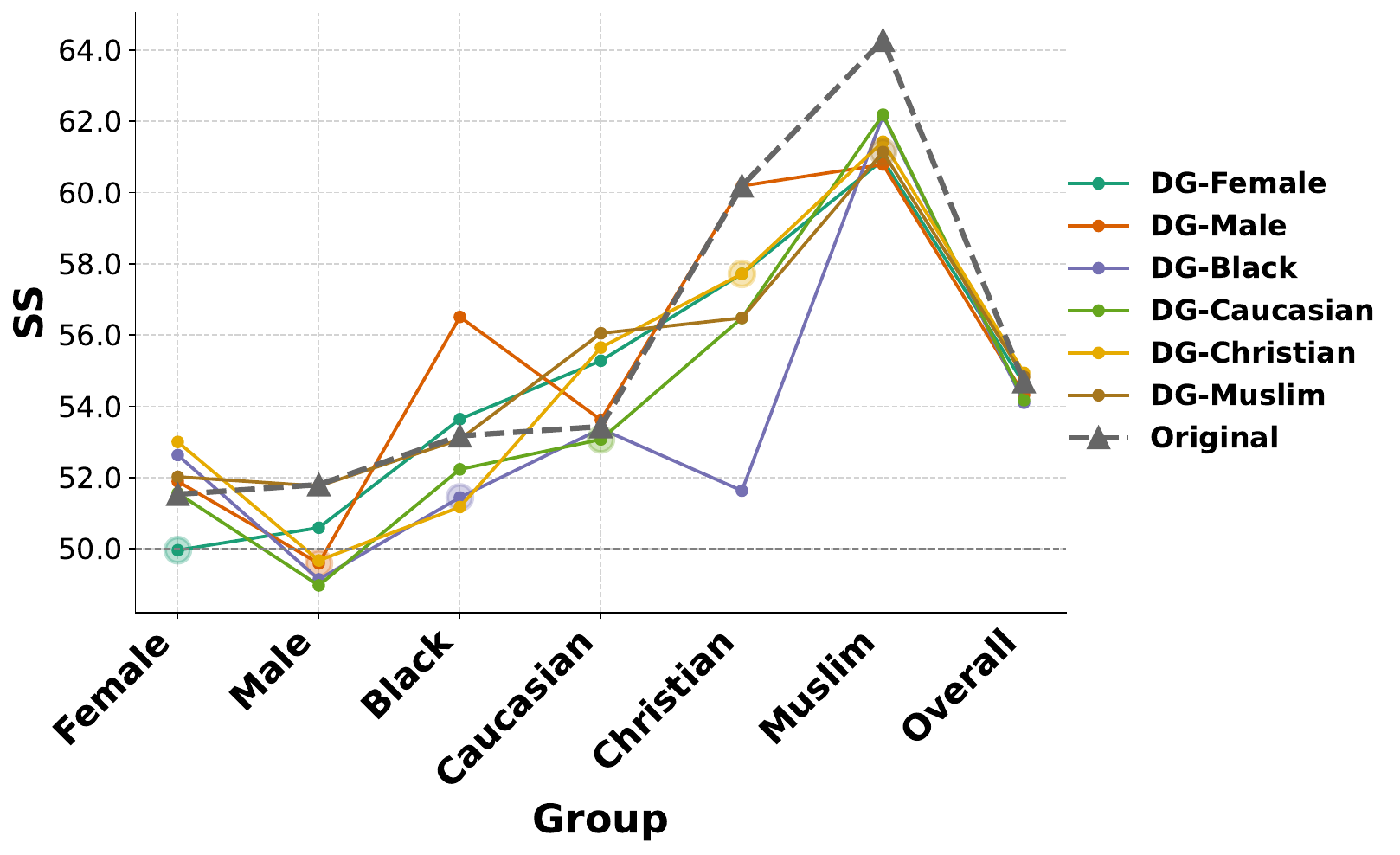}
        \caption{TinyBERT-DG}
        \label{fig:pre-full:tinybert-dg}
    \end{subfigure}
    ~
    \begin{subfigure}[b]{.32\linewidth}
        \includegraphics[width=\linewidth]{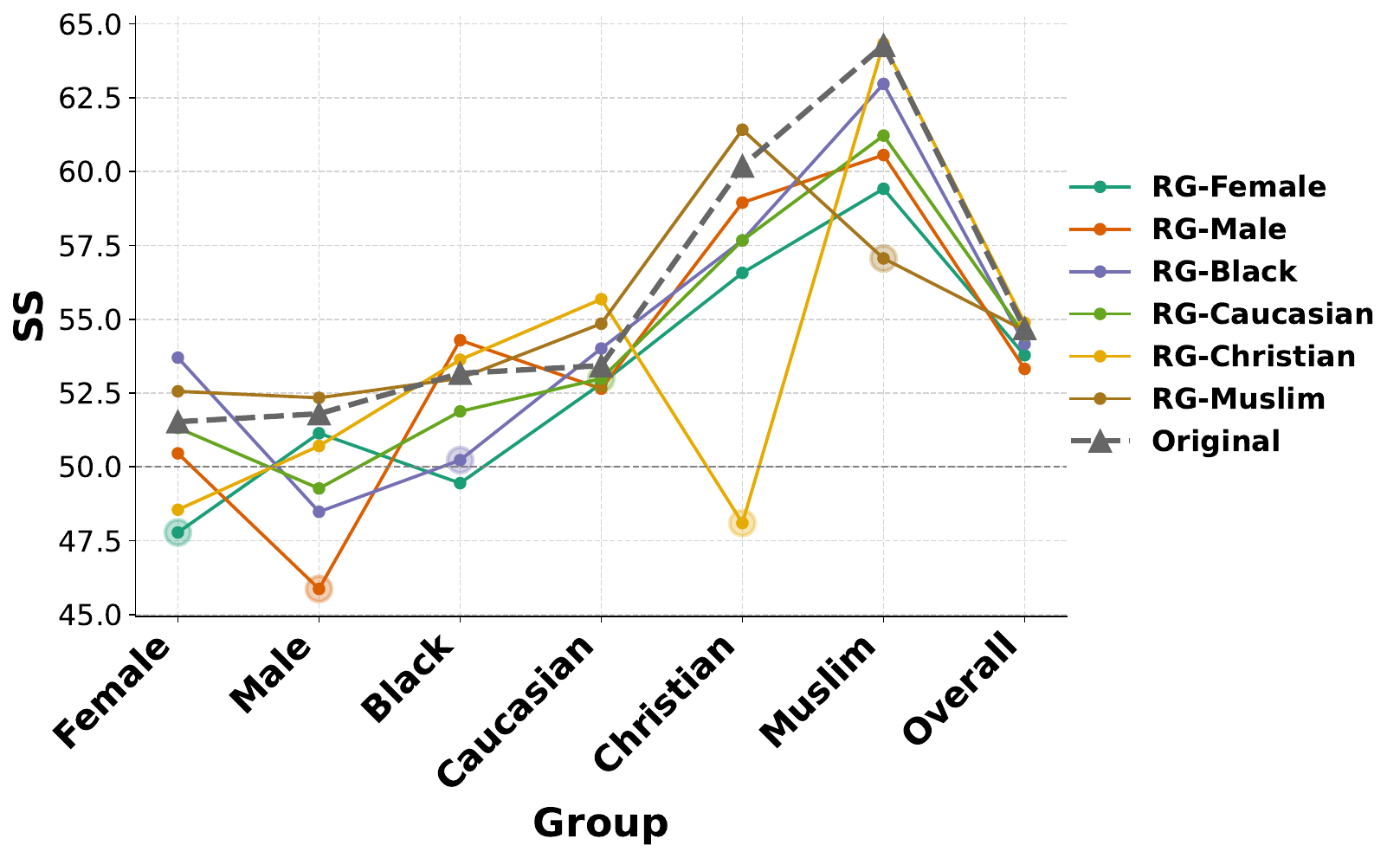}
        \caption{TinyBERT-RG}
        \label{fig:pre-full:tinybert-rg}
    \end{subfigure}
    ~
    \begin{subfigure}[b]{.32\linewidth}
        \includegraphics[width=\linewidth]{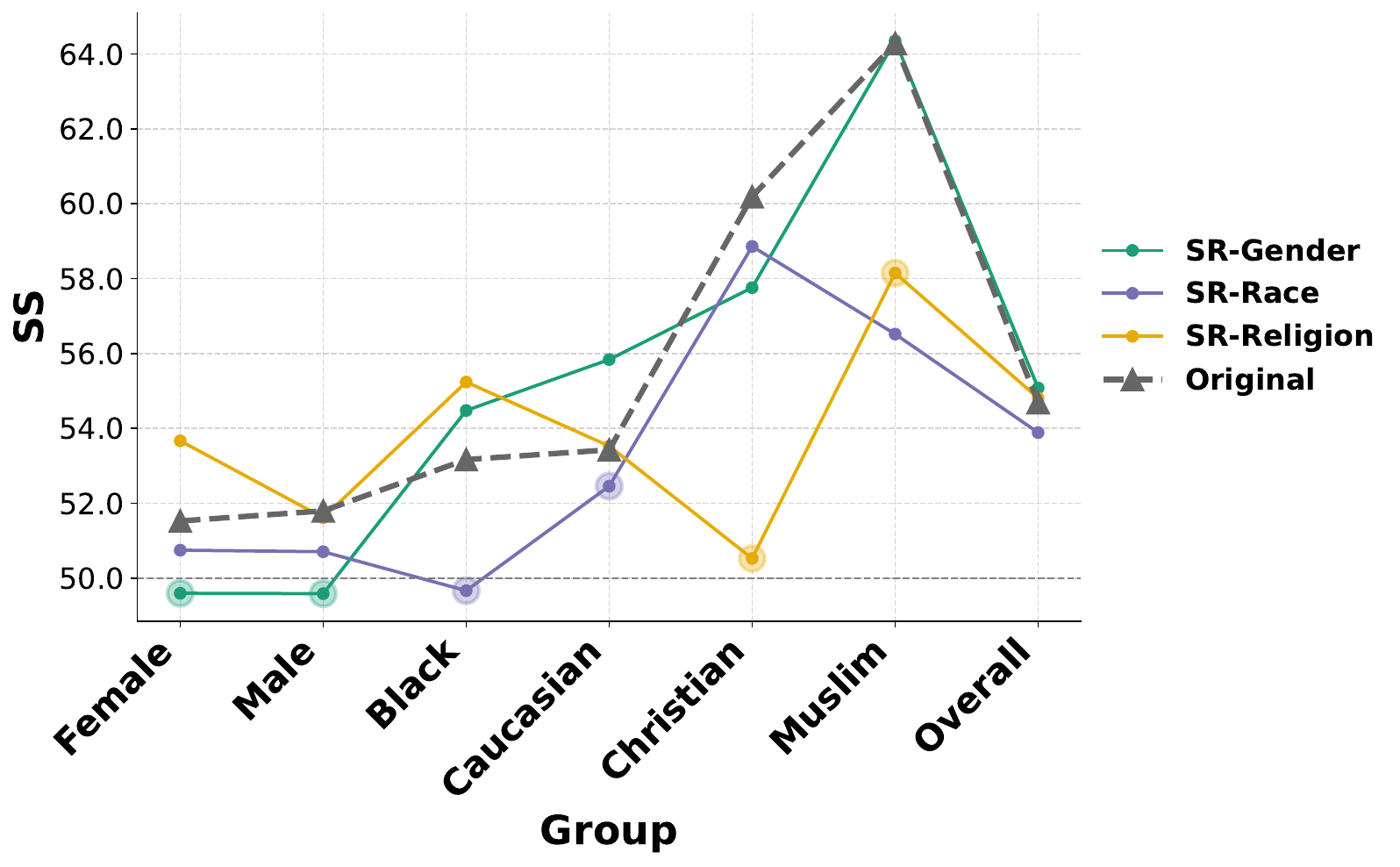}
        \caption{TinyBERT-SR}
        \label{fig:pre-full:tinybert-sr}
    \end{subfigure}

    \begin{subfigure}[b]{.32\linewidth}
        \includegraphics[width=\linewidth]{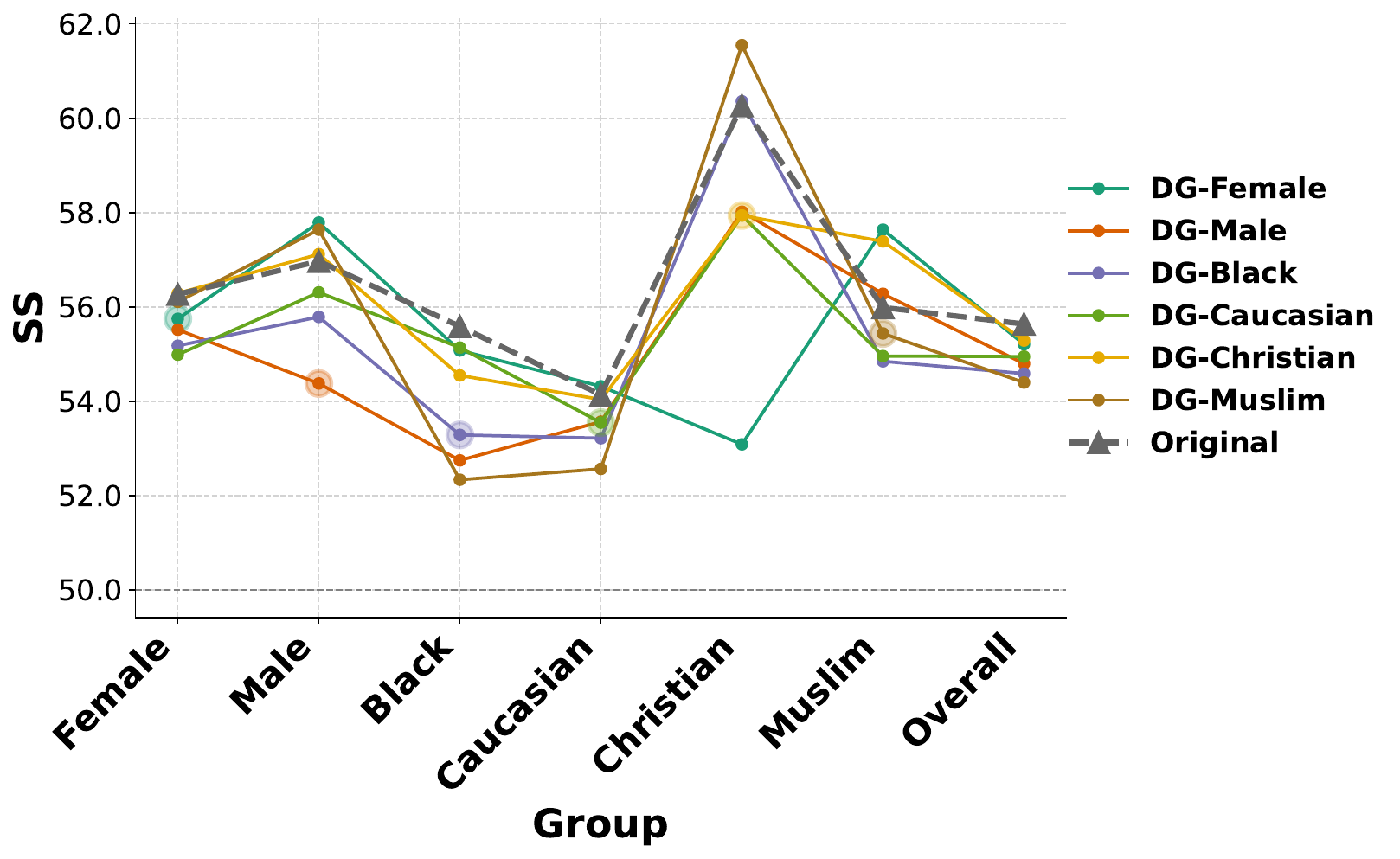}
        \caption{GPT-2-DG}
        \label{fig:pre-full:gpt2-dg}
    \end{subfigure}
    ~
    \begin{subfigure}[b]{.32\linewidth}
        \includegraphics[width=\linewidth]{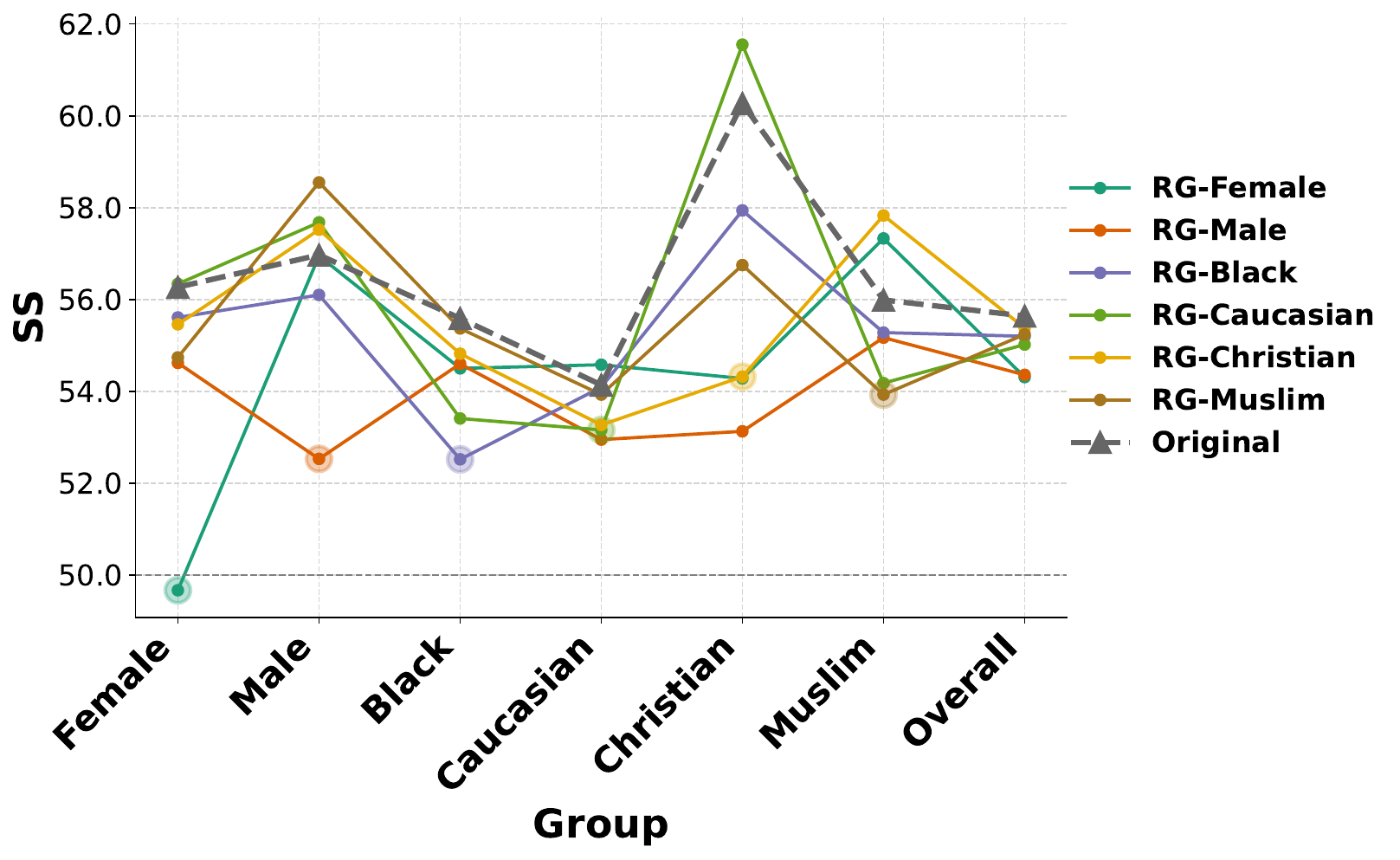}
        \caption{GPT-2-RG}
        \label{fig:pre-full:gpt2-rg}
    \end{subfigure}
    ~
    \begin{subfigure}[b]{.32\linewidth}
        \includegraphics[width=\linewidth]{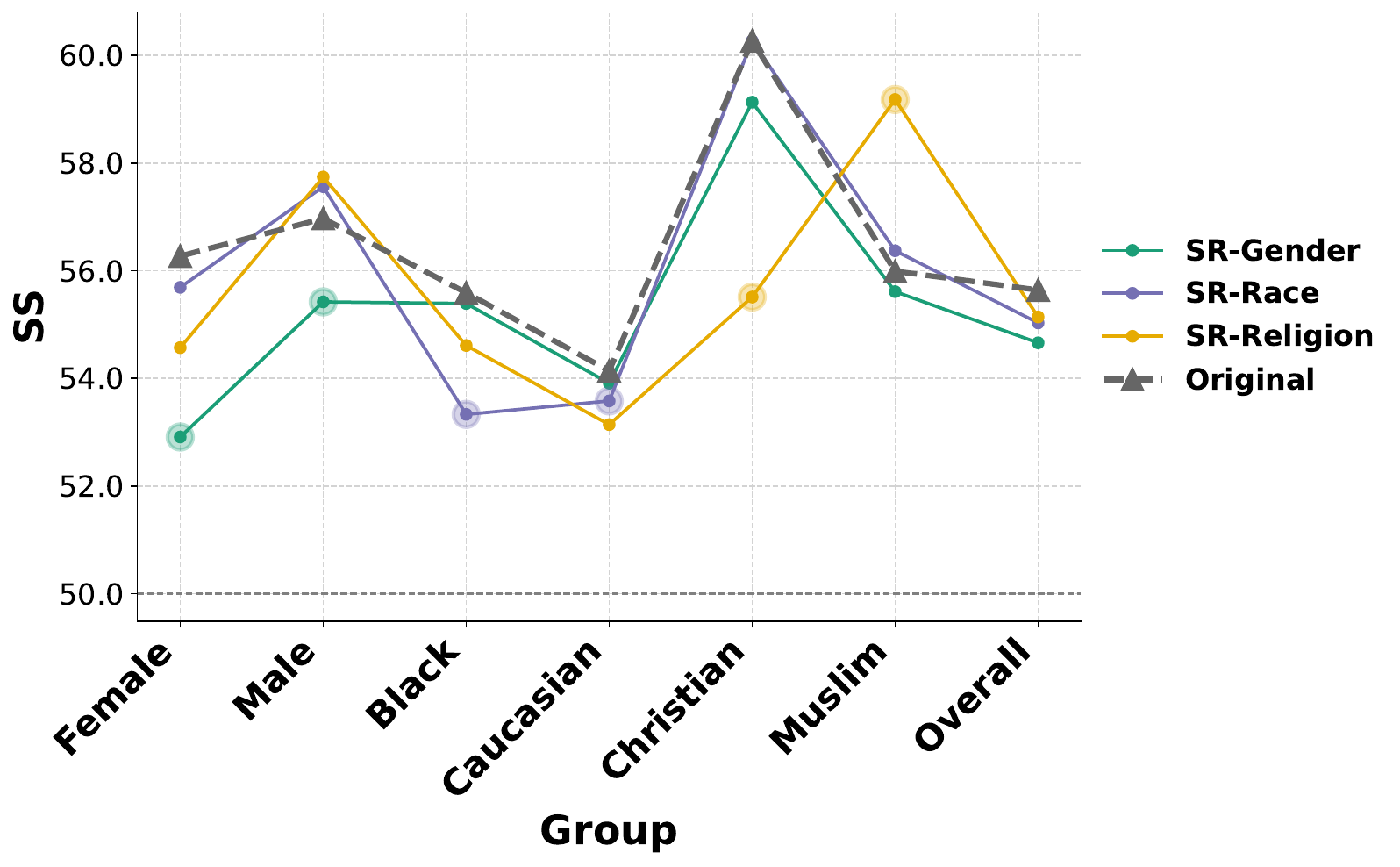}
        \caption{GPT-2-SR}
        \label{fig:pre-full:gpt2-sr}
    \end{subfigure}
    \caption{StereoSet stereotype scores (SS; 50 is neutral, closer to 50 indicates \emph{less} stereotyping/anti-stereotyping) for models pre-trained on debiased Wikipedia under three preprocessing strategies. Dashed lines mark the original model; Larger semi-transparent markers indicate the target group used for data cleaning. Full numeric results appear in Appendix Tables \ref{tbl:stereoset-full-pre-dg}-\ref{tbl:stereoset-full-pre-sr} and \ref{tbl:gpt2-stereoset-full-pre-dg}-\ref{tbl:gpt2-stereoset-full-pre-sr}.}\label{fig:pre-full}
    \vspace{-3mm}
\end{figure*}

\begin{figure*}[!h]
    \centering
    \begin{subfigure}[b]{.32\linewidth}
        \includegraphics[width=\linewidth]{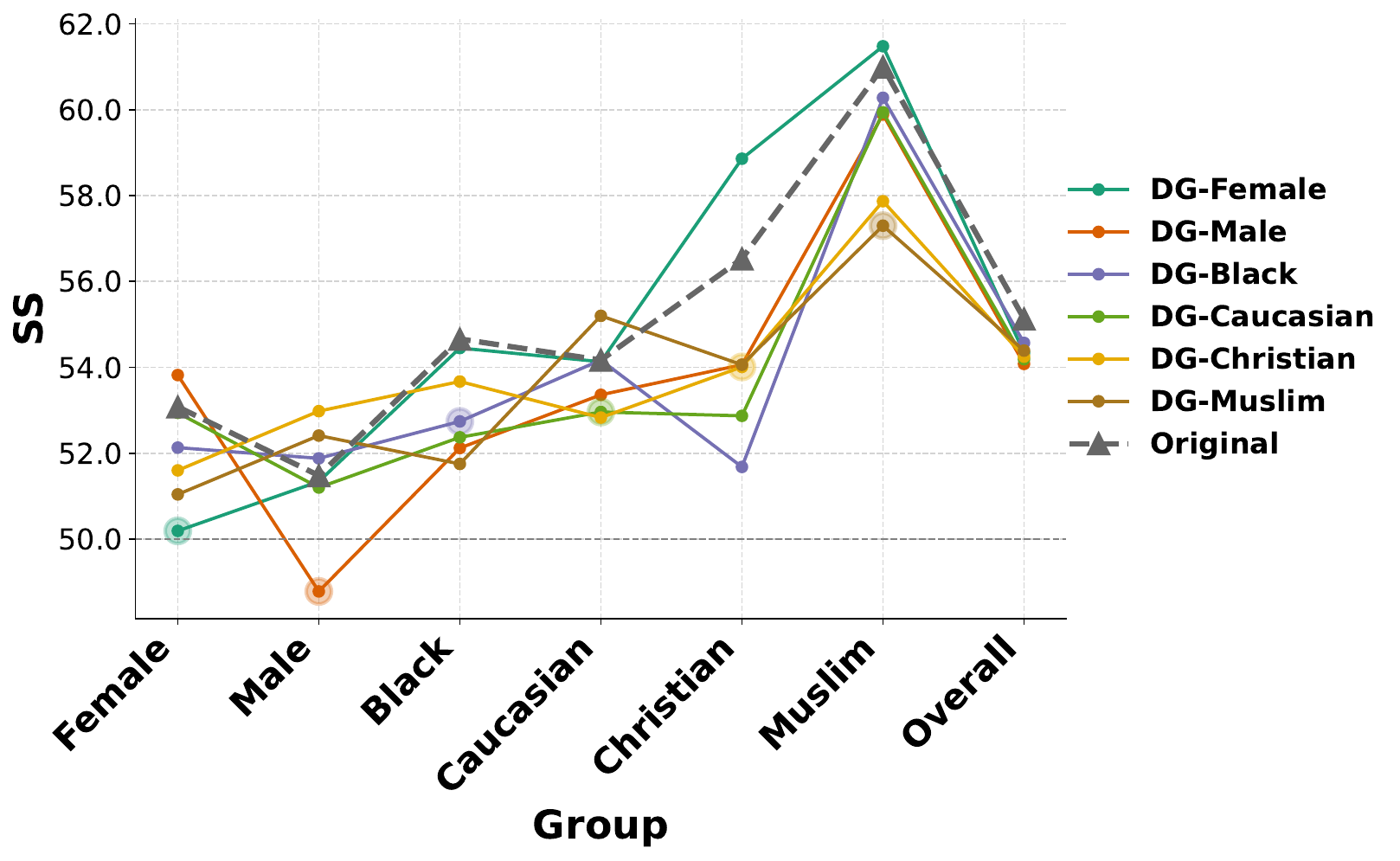}
        \caption{TinyBERT-DG}
        \label{fig:post-full:tinybert-dg}
    \end{subfigure}
    ~
    \begin{subfigure}[b]{.32\linewidth}
        \includegraphics[width=\linewidth]{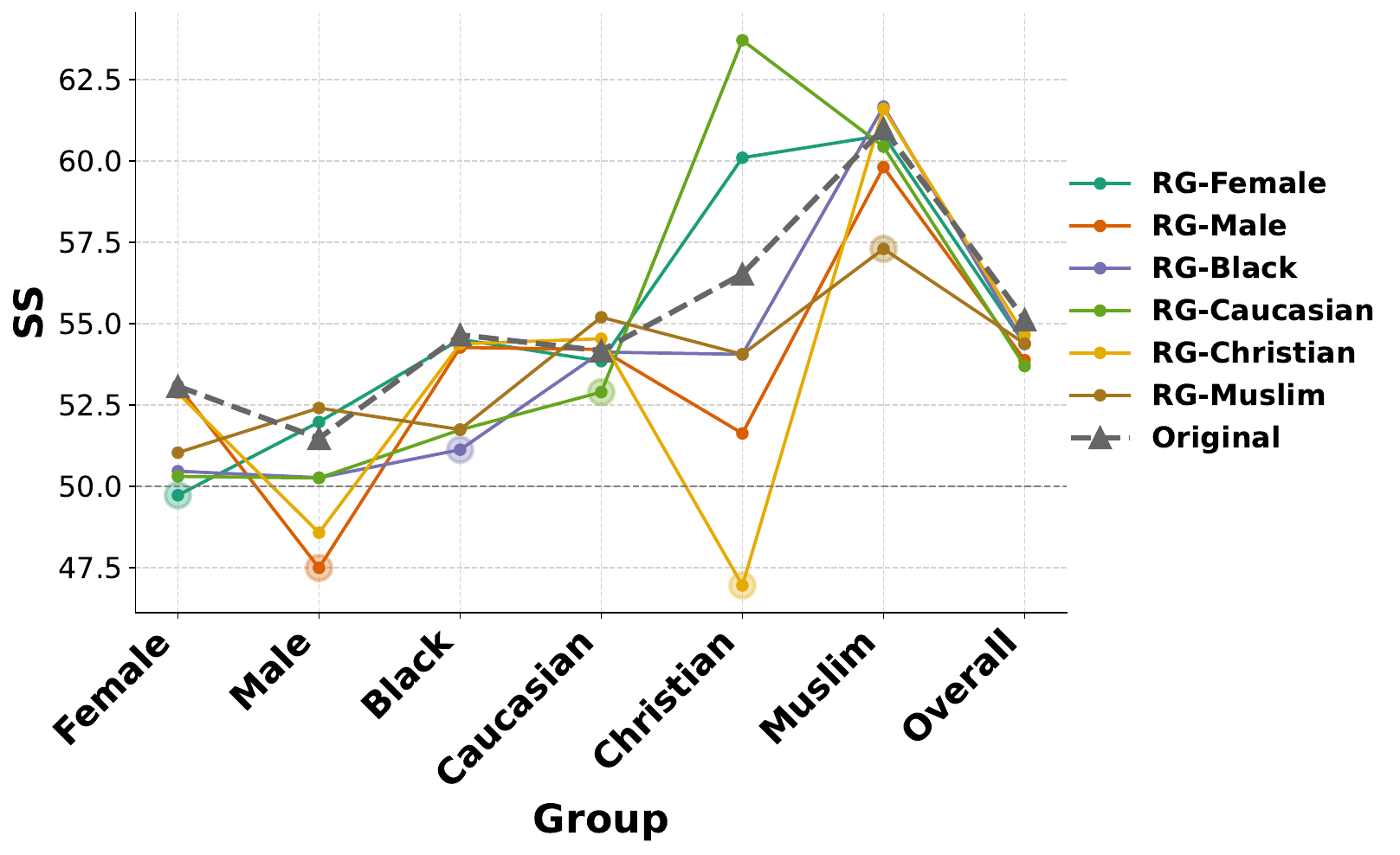}
        \caption{TinyBERT-RG}
        \label{fig:post-full:tinybert-rg}
    \end{subfigure}
    ~
    \begin{subfigure}[b]{.32\linewidth}
        \includegraphics[width=\linewidth]{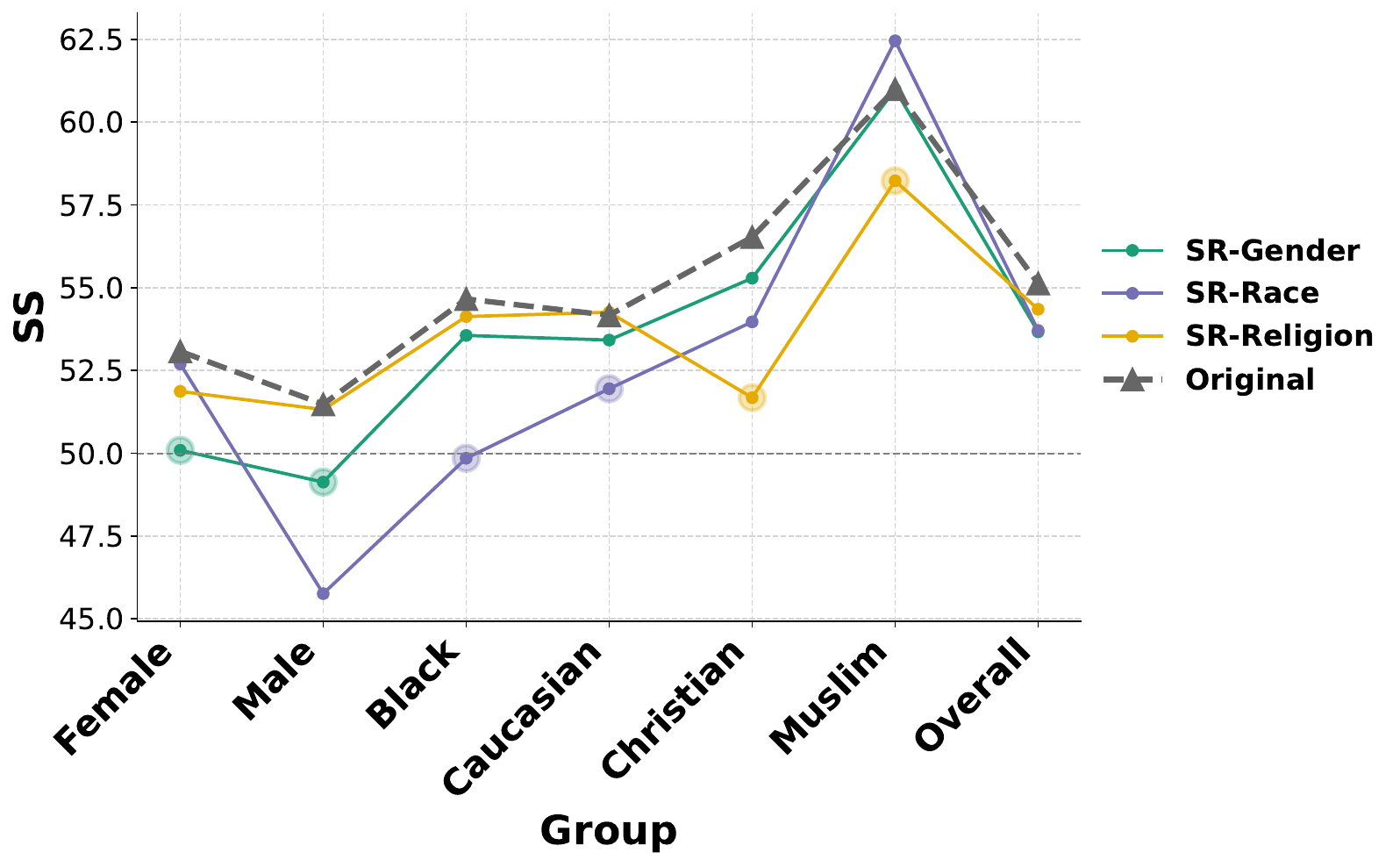}
        \caption{TinyBERT-SR}
        \label{fig:post-full:tinybert-sr}
    \end{subfigure}

    \begin{subfigure}[b]{.32\linewidth}
        \includegraphics[width=\linewidth]{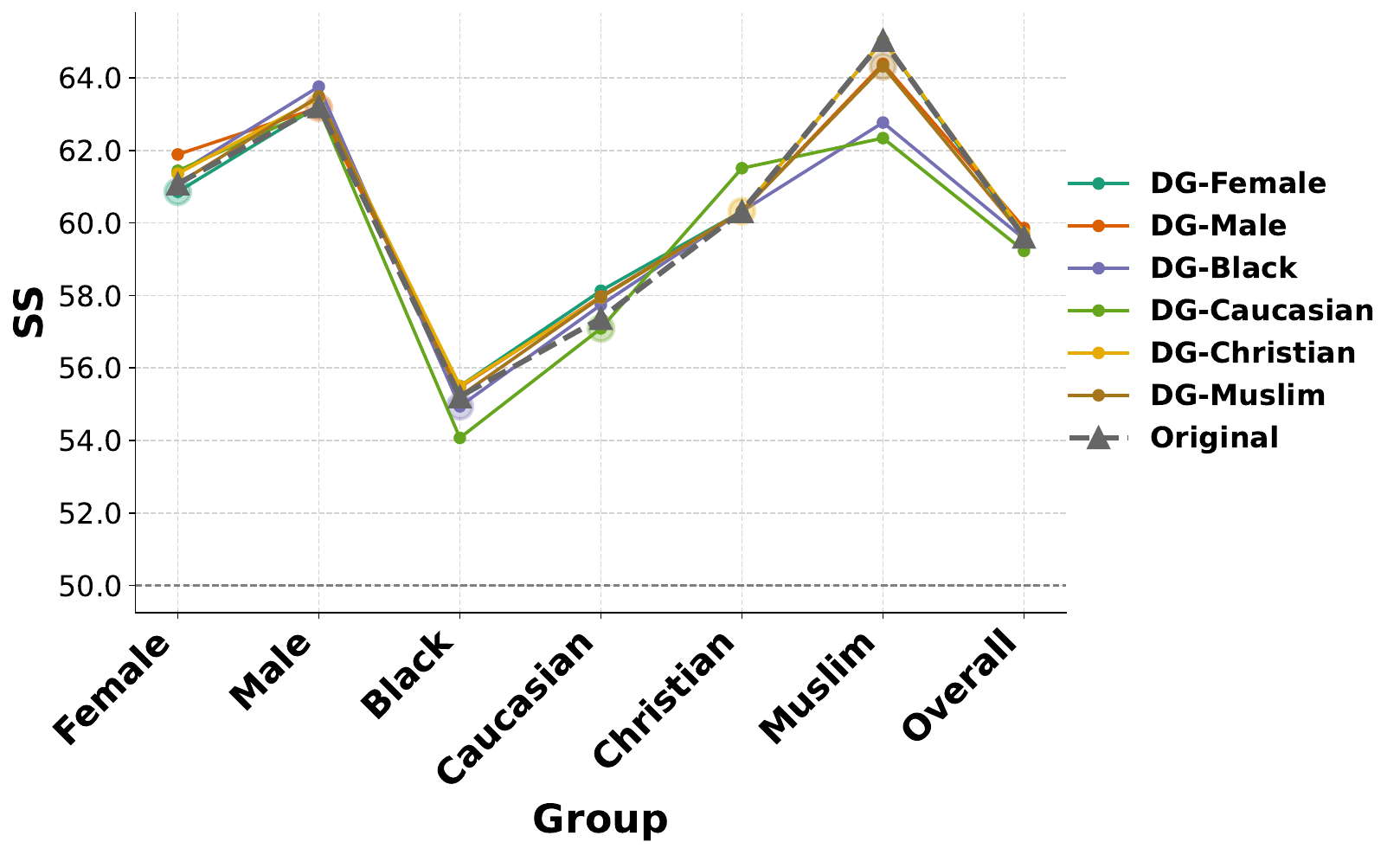}
        \caption{GPT-2-DG}
        \label{fig:post-full:gpt2-dg}
    \end{subfigure}
    ~
    \begin{subfigure}[b]{.32\linewidth}
        \includegraphics[width=\linewidth]{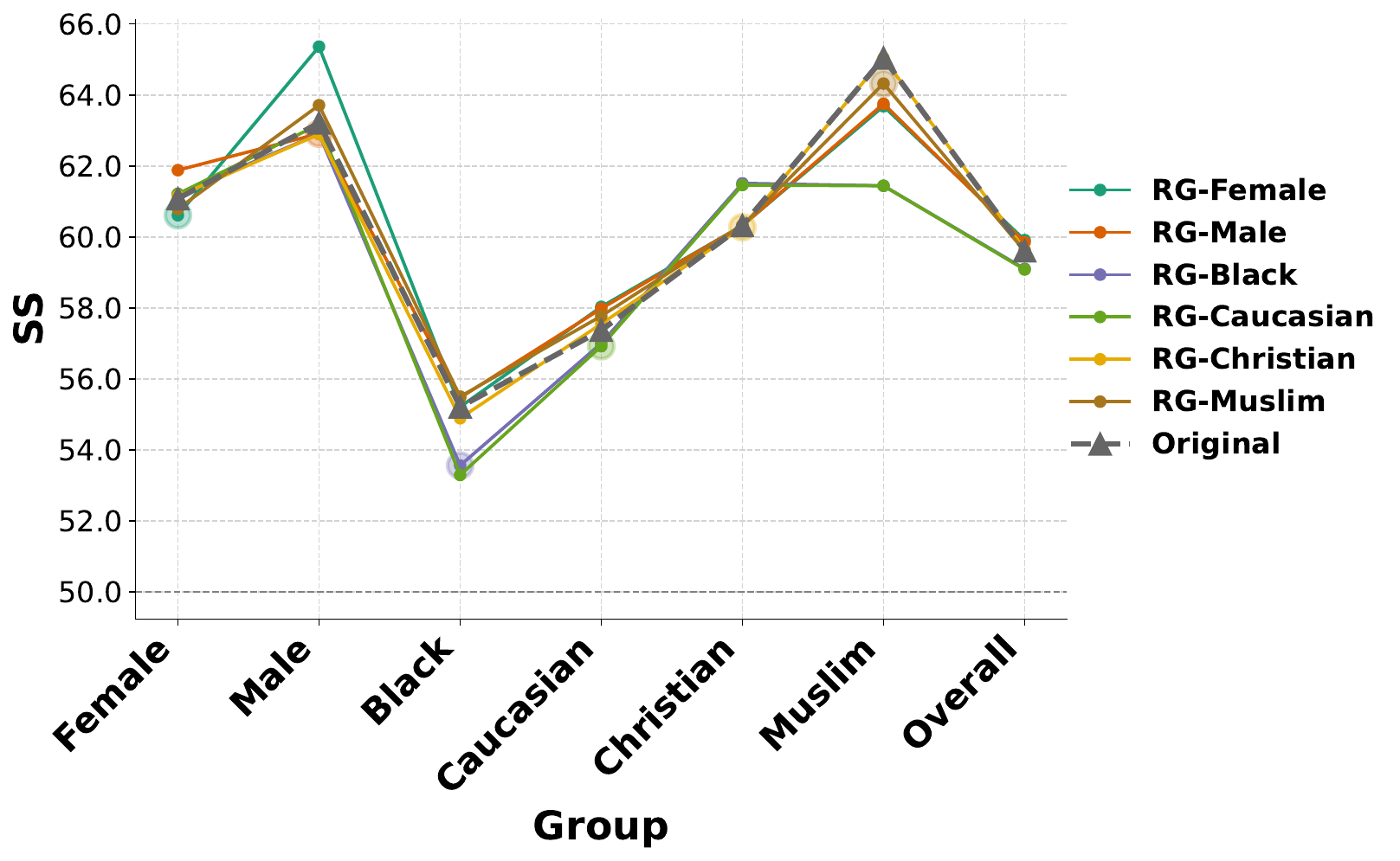}
        \caption{GPT-2-RG}
        \label{fig:post-full:gpt2-rg}
    \end{subfigure}
    ~
    \begin{subfigure}[b]{.32\linewidth}
        \includegraphics[width=\linewidth]{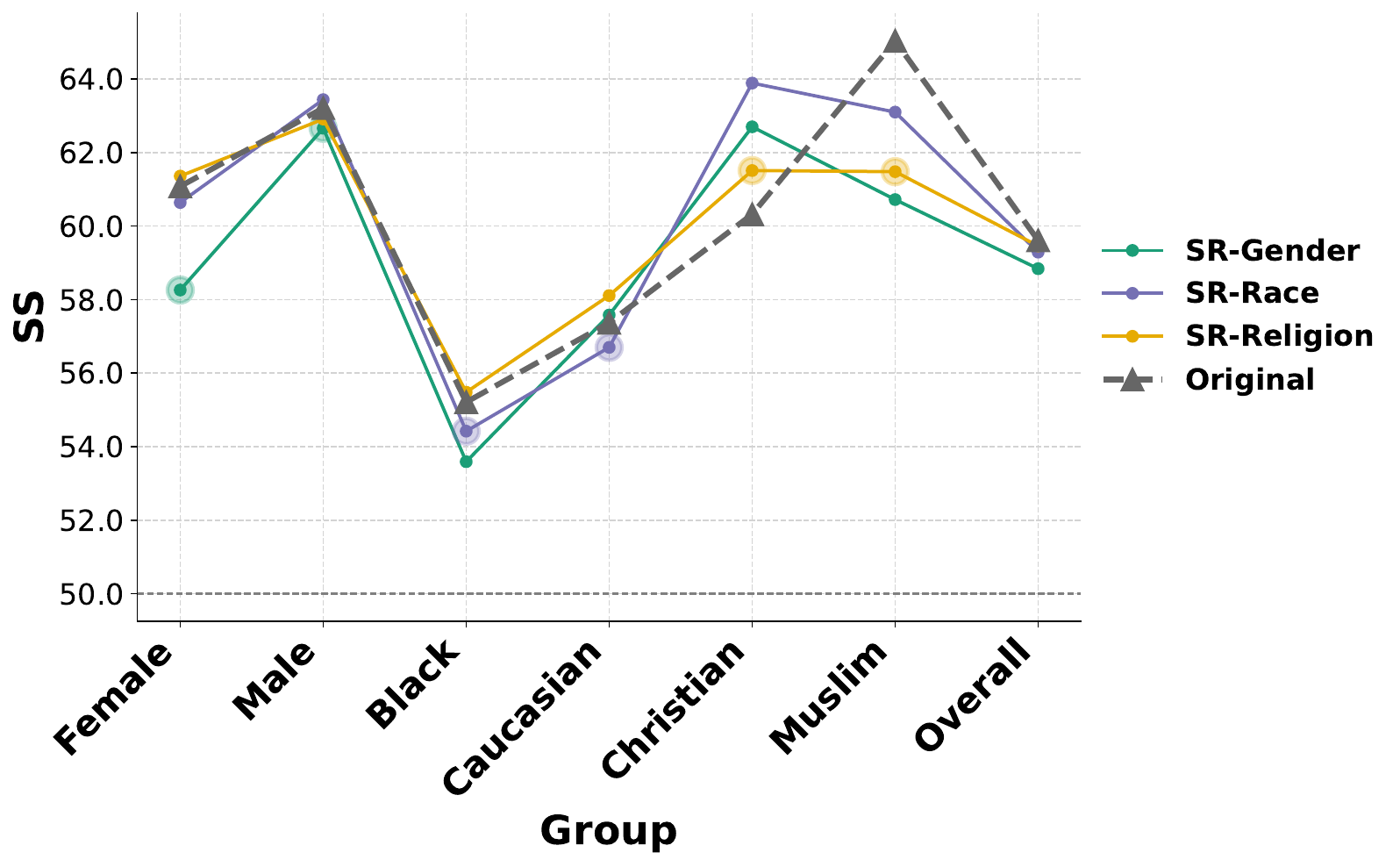}
        \caption{GPT-2-SR}
        \label{fig:post-full:gpt2-sr}
    \end{subfigure}
       \caption{StereoSet stereotype (SS) for models \emph{post-trained} on debiased Wikipedia. Full numeric results appear in Appendix Tables \ref{tbl:stereoset-full-post-dg}-\ref{tbl:stereoset-full-post-sr} and \ref{tbl:gpt2-stereoset-full-post-dg}-\ref{tbl:gpt2-stereoset-full-post-sr}.}
    \label{fig:post-full}
    \vspace{-10pt}
\end{figure*}

We first evaluate the extent to which preprocessing-based methods reduce stereotype scores for the targeted groups. We evaluate the effects of three preprocessing strategies (DG, RG, SR) under both pre-training and post-training of TinyBERT and GPT-2 on the full Wikipedia snapshot and report StereoSet's stereotype score (SS), language modeling score (LMS), and iCAT together with per-group SS, following \citet{nadeem-etal-2021-stereoset}. We also report CrowS-Pairs results for completeness (Appendix~\ref{appendix:stereo-eval:crows}). 

\textbf{Pre-training.} As shown in Fig.~\ref{fig:pre-full}, SS for the \emph{target} group typically decreases relative to the baseline models, moving toward 50 for both TinyBERT and GPT-2, with some limited exceptions (e.g., GPT-2 under SR-religion shows increased SS for Muslims). Similar trends are observed in the CrowS-Pairs benchmark (Appendix \ref{appendix:stereo-eval:crows}). Aggregating across groups, overall SS often decreases while LMS remains comparable, yielding iCAT that is similar to or higher than the base models. 

\textbf{Post-training.} Post-training generally yields larger reductions in target-group SS than pre-training (Fig.~\ref{fig:post-full}), again with broadly stable LMS, yielding higher iCAT scores overall. Detailed StereoSet and CrowS-Pairs results are provided in Appendices~\ref{appendix:stereo-eval:stereoset:post} and~\ref{appendix:stereo-eval:crows:post}.

\textbf{Data scale.} Using $5\%$ of Wikipedia, both pre-training and post-training still consistently reduce SS for targeted groups. These improvements occur alongside stable or slightly enhanced LMS and iCAT scores. These results are shown in Appendix \ref{appendix:stereo-results}.

Taken together, these results show that preprocessing-based debiasing can reduce stereotype scores for targeted groups in a data-efficient manner. However, as we show in the next section, these improvements often coincide with unintended side effects in non-target groups.

\section{Data Debiasing Incurs Side Effects} 
\label{sct:side-effects}

Alongside the targeted improvements documented in Section~\ref{sct:exp-positive}, we observe \emph{side effects}: stereotype scores (SS) for \emph{non-target} groups sometimes move \emph{away from 50}. These shifts vary by preprocessing strategy and training stage (pre- vs.\ post-training), and they can cross stereotype categories. We therefore emphasize directional changes relative to the corresponding base model and highlight representative patterns below.

\subsection{Stereotype Shifts Within Categories}
For each category (gender, race, religion), we evaluate two demographic groups to enable within-category comparisons. While debiasing a given group often moves its SS toward 50, the \emph{other} group in the same category does not behave uniformly.

For example, pre-training TinyBERT on the full Wikipedia corpus debiased for females (DG-female) reduces the model's stereotypes toward males (green line in Figure~\ref{fig:pre-full:tinybert-dg}), whereas debiasing for males (DG-male) increases stereotypes toward females (red line in Figure~\ref{fig:pre-full:tinybert-dg}). Post-training shows analogous behavior: post-training on DG-male increases stereotypes toward females (red line in Figure~\ref{fig:post-full:tinybert-dg}), and post-training on RG-female increases stereotypes toward males (green line in Figure~\ref{fig:post-full:tinybert-rg}).

Interestingly, these spillovers depend on the debiasing method, which complicates a simple distributional explanation. TinyBERT pre- and post-trained on DG-male both show increased stereotypes toward females (Figures~\ref{fig:pre-full:tinybert-dg} and~\ref{fig:post-full:tinybert-dg}), whereas models trained on RG-male exhibit inconsistent patterns across pre- versus post-training (Figures~\ref{fig:pre-full:tinybert-rg} and~\ref{fig:post-full:tinybert-rg}).

Unintended effects also extend beyond directly opposing groups. For instance, pre-training on RG-Muslim increases stereotypes toward Christians (Figure~\ref{fig:pre-full:tinybert-rg}), even though the two groups are not strict opposites in the StereoSet framework.

Taken together, preprocessing-based debiasing induces non-uniform and hard-to-predict shifts. The observed changes are not readily explained by altered data distributions and appear non-systematic across groups and training regimes. This raises a practical concern: \emph{removing stereotypes documented in existing benchmarks may not mitigate bias, but instead redirect it toward other groups, including groups not represented in those benchmarks.}

\subsection{Stereotype Variability Across Categories}
Cross-category side effects are often more pronounced than within-category shifts and can reverse across training stages. For instance, the TinyBERT model pre-trained on the preprocessed Wikipedia corpus under the RG-female setting shows a consistent reduction in stereotype levels across all 6 groups evaluated in our study (Figure \ref{fig:pre-full:tinybert-rg}), whereas post-training on the same RG–female corpus increases Christian SS (Figure~\ref{fig:post-full:tinybert-rg}).  

\begin{figure}[t]
    \centering
    \includegraphics[width=1\linewidth]{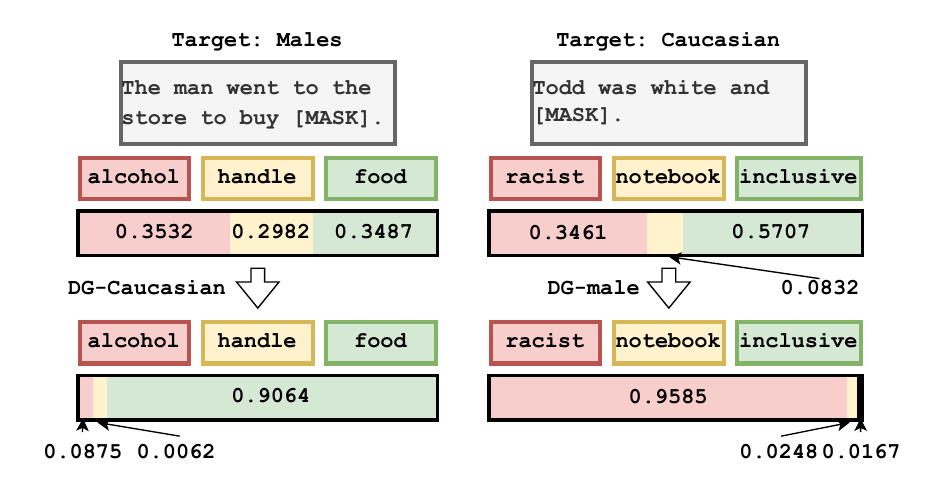}
    \caption{Example of cross-group stereotype shifts induced by preprocessing-based mitigation. Pre-training TinyBERT on DG-Caucasian data leads to anti-stereotypical behavior toward males (e.g., preferring “food” over “alcohol”), while DG-male induces stronger stereotypical behavior toward Caucasians (e.g., preferring “racist” over “inclusive”). These asymmetric shifts, reflected in token-level probabilities, highlight the unpredictability of side effects across groups.}
    \label{fig:side-effect-exp}
    \vspace{-10pt}
\end{figure}

These cross-category side effects are not symmetric either. Comparing 2 TinyBERT models pre-trained on the full Wikipedia corpus cleaned under RG-female and RG-Christian settings, we observe that the former displays elevated stereotype levels toward Christians, whereas the latter actually reduces stereotypes toward females (Figure \ref{fig:pre-full:tinybert-rg}).
We observed similar trends when manually analyzing the models' word preferences before and after stereotype-mitigation training. Figure~\ref{fig:side-effect-exp} provides a concrete example. 

As shown in Figure~\ref{fig:side-effect-exp}, pre-training TinyBERT on DG-cleaned data for the Caucasian group shifts the model's behavior from exhibiting mild stereotypical tendencies to predominantly generating anti-stereotypical completions. For the sentence ``That man went to the store to buy [BLANK],''
the model prefers ``food'' over ``alcohol,'' countering the stereotype that ``men are heavy drinkers.'' However, this effect is not symmetric. When TinyBERT is instead pre-trained on DG-male data, it becomes highly stereotypical when presented with the sentence ``Todd was white and [BLANK],'' assigning a much higher probability to ``racist'' than to neutral alternatives such as ``inclusive.'' 

These effects are accompanied by substantial shifts in token-level probabilities, as shown in Figure~\ref{fig:side-effect-exp}. The lack of a consistent relationship between data interventions and behavioral outcomes, together with the observed asymmetry, highlights the unpredictability of stereotype-mitigation side effects. This unpredictability complicates the assessment of mitigation effectiveness and poses significant challenges for the safe and reliable deployment of PLMs.

Even more notably, the stereotype that ``men are heavy drinkers'' does not appear in StereoSet, the benchmark used to guide the debiasing of the Wikipedia corpus. This raises additional concerns about preprocessing-based debiasing approaches, as they may inadvertently introduce or amplify stereotypes that are not covered by existing benchmarks. As a result, it becomes difficult or impossible to fully assess the potential harms introduced by such debiasing, given that the evaluation framework lacks coverage of these emergent or untracked biases.

These side effects happen to models trained using other preprocessing settings as well.
For example, the model pre-trained on DG-male shows lower-than-usual stereotype levels toward Christians while pre-training the model on DG-Christian leads to much higher stereotype levels toward males (Figure \ref{fig:pre-full:tinybert-dg}).
This observation counters the potential explanation that side effects are the results of the stereotype/anti-stereotype content distributions being affected by text removal, since if stereotypical content toward males overlaps with stereotypical content toward Christians, DG for Christians should also reduce the model's stereotype levels on the male group.
Hence, these results cannot be fully explained by changes in stereotype/anti-stereotype content distributions alone, suggesting that additional mechanisms may contribute to the observed side effects. Importantly, such side effects can undermine the effectiveness and reliability of preprocessing-based stereotype mitigation approaches by introducing unintended shifts in non-target groups.

The stereotype evaluation results of models pre- or post-trained on the SR-cleaned corpora further consolidate the randomness of such side effects.
Theoretically, switching the references to a pair of groups would by no means affect the ratio of stereotypical contents toward other groups in a corpus.
Yet, we noted that the TinyBERT model pre-trained on the SR-cleaned corpus for racial stereotypes (affecting Black people and Caucasian people) shows a higher stereotype level toward Christians and a higher anti-stereotype level toward males (Figure \ref{fig:pre-full:tinybert-sr}).
Leveraging the same data, the model post-trained on the SR-cleaned data for racial stereotypes shows harsher stereotypes toward Muslims while a lower stereotype level toward Christians (Figure \ref{fig:post-full:tinybert-sr}).
Due to the existence of random side effects with high frequency in our experiments, it raises concerns about the reliability, effectiveness, and robustness of preprocessing-based stereotype mitigation approaches.

\subsection{Model Agnosticity of Side Effects} \label{sct:discussion-model}
The experimental results on GPT-2 demonstrate that the side effects of preprocessing-based stereotype mitigation extend beyond small encoder-only models trained with the MLM objective.
As shown in Figure \ref{fig:pre-full:gpt2-dg}, pre-training GPT-2 on DG-female data amplifies stereotypes toward males, whereas DG-male consistently reduces SS for both gender groups. Unexpectedly, both DG-male and DG-female increase stereotypes toward Muslims. Removing female mentions entirely (RG-female) minimizes negative impacts on male-oriented stereotypes but continues to strengthen Muslim-oriented stereotypes (Figure \ref{fig:pre-full:gpt2-rg}). Surprisingly, switching male and female references (SR-Gender) mitigates stereotypes across all 6 groups, resulting in overall lower stereotype levels on StereoSet (Figure \ref{fig:pre-full:gpt2-sr}).
Similar unpredictable and non-systematic stereotype shifts appear in other models, indicating that stereotype expression is a confounding issue across different types of LLMs. Although smaller in magnitude, these side effects persist in GPT-2 post-training experiments (Figures \ref{fig:post-full:gpt2-dg}–\ref{fig:post-full:gpt2-sr}).

These findings confirm that the unintended consequences of preprocessing-based stereotype mitigation are not confined to small encoder models. Rather, they may undermine the reliability of debiasing across a wide range of PLMs, highlighting the need for more robust evaluation frameworks and mitigation strategies that explicitly account for cross-group side effects.

\section{Side Effects Beyond Semantics}
\begin{figure}[t]
    \centering
    \includegraphics[width=1\linewidth]{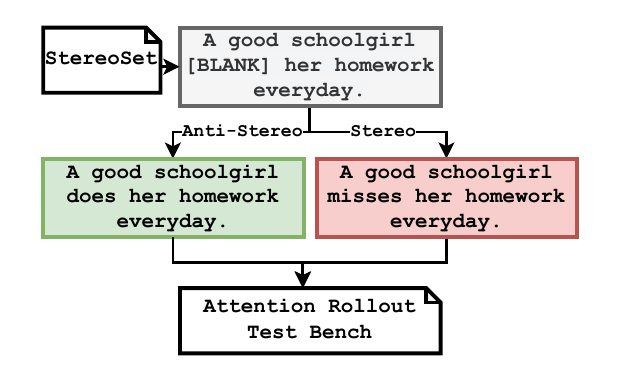}
    \caption{Curation process for the attention-rollout test bench derived from StereoSet instances.}
    \label{fig:attention-rollout-prep}
    \vspace{-10pt}
\end{figure}

\begin{figure*}[t]
\centering
    \begin{subfigure}[b]{\linewidth}
        \includegraphics[width=\linewidth]{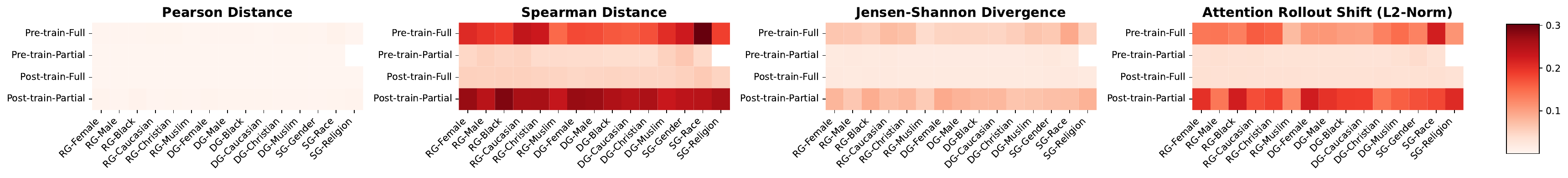}
        \caption{TinyBERT}
    \end{subfigure}

    \begin{subfigure}[b]{\linewidth}
        \includegraphics[width=\linewidth]{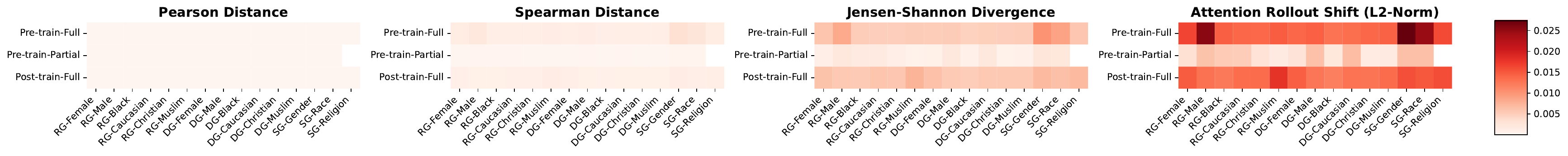}
        \caption{GPT-2}
    \end{subfigure}
   \caption{Distances/divergences between attention-rollout distributions of stereotype-mitigated models and their base counterparts, plus L2 norms of rollout shifts (darker indicates larger differences). All differences shown are statistically significant (p < 0.001).}
    \label{fig:rollout-analysis}
    \vspace{-10pt}
\end{figure*}

Removing content from LLM training corpora can alter models’ semantic representations due to attention allocation changes. To assess whether semantic shifts might still contribute, we compute attention rollout~\cite{attention-rollout} for each model on StereoSet inputs and compare rollout values between stereotype-mitigated models and their corresponding base models.

Attention rollout estimates how information from original input tokens contributes to representations at a chosen layer. In contrast to raw, per-layer attention, which shows only same-layer interactions among already-mixed embeddings, rollout aggregates heads, accounts for residual connections, and multiplies attention matrices across layers to propagate the influence. The resulting joint map provides a row-normalized distribution approximating relative influence along all attention paths. In our setting, rollout identifies the input tokens that most strongly affect internal representations and enables direct, cross-model comparisons of attribution patterns to reveal how debiasing affects attention.

We use StereoSet data to ensure that all test text directly concerns the minority groups under study, avoiding dilution from unrelated content. For data preparation, each StereoSet instance is converted into 2 sentences (one stereotypical and one anti-stereotypical), as illustrated in Figure \ref{fig:attention-rollout-prep}. For completeness, we compare each stereotype-mitigated model with its base counterpart using Pearson distance (PD), Spearman distance (SD), Jensen–Shannon divergence (JSD), and the L2-norm of attention shift (AS) on rollout distributions. Among these metrics, PD and SD capture distributional correlation differences, AS quantifies absolute changes in attention allocation, and JSD measures divergence between normalized attention distributions \cite{jsd-justification-1,jsd-justification-2}. Results are shown as heatmaps in Figure \ref{fig:rollout-analysis}.

Overall, the attention rollout patterns remain highly consistent between stereotype-mitigated models and their base counterparts, with very low maximum distances across all metrics (PD: 0.0061; SD: 0.3029; JSD: 0.0925; AS: 0.2235). This consistency holds at both the distribution level (PD, SD, JSD) and the magnitude level (AS), and it is robust across model sizes, from smaller architectures such as TinyBERT to larger ones like GPT-2.

Finer-grained, group-specific analyses reach the same conclusion, even for models that show pronounced side effects on stereotype metrics. For example, the TinyBERT model pre-trained under the DG-male setting increases stereotype scores for women and, more strongly, for Black people (Figure~\ref{fig:pre-full:tinybert-dg}), yet rollout shifts remain small (PD: 0.001279, SD: 0.162548, JSD: 0.050337, AS: 0.103921 for women; PD: 0.001300, SD: 0.157040, JSD: 0.049205, AS: 0.101581 for Black people). Similarly, GPT-2 pre-trained under the DG-female setting lowers stereotype scores for Christians and increases them for Muslims, while rollout shifts remain comparably small (PD: 0.000020, SD: 0.001176, JSD: 0.004793, AS: 0.013177 for Christian; PD: 0.000020, SD: 0.001015, JSD: 0.004882, AS: 0.012909 for Muslim).

Taken together, these results suggest that the observed side effects of stereotype mitigation are not well explained by changes in how models semantically route information, as captured by attention rollout. While we note that our representation-level analysis only provides surface-level insights into each model, it serves as a useful diagnostic probe to demonstrate the unclear origin of the side effects we observe.

Their underlying mechanism remains uncertain, motivating methods that probe alternative causal pathways and a reconsideration of stereotype benchmarking practices to ensure faithful, side-effect-aware safety evaluation of LLMs. We note that identifying the precise layers, cues, or circuits responsible for stereotype redistribution is beyond the scope of this work.

\section{Discussion}
We additionally test the robustness of our findings across models of different training data sizes (Section \ref{sct:discussion-data-size}), and for massive LLMs (Section \ref{sct:discussion-llama}).

\subsection{Data Size Agnosticity of Side Effects} \label{sct:discussion-data-size}
According to our experimental results, the DG, RG, and SR approaches using a tiny subset of the Wikipedia corpus (5\%) still lead to less biased models for the target groups (Appendix \ref{appendix:stereo-results}).
This further highlights the sensitivity of models to preprocessing-based stereotype mitigation approaches.

Alongside the benefits, the side effects of stereotype mitigation still exist and are unpredictable.
For example, stereotypes on the Caucasian group become more pronounced when the TinyBERT model is pre-trained on the SR-cleaned data for gender stereotypes (Table \ref{tbl:stereoset-subset-pre-sr}).
Gender stereotypes toward both groups get worse in the model post-trained on the SR-cleaned data for racial stereotypes (Table \ref{tbl:stereoset-partial-post-sr}) while they are reduced in the pre-trained model on the same data (Table \ref{tbl:stereoset-subset-pre-sr}).

From a finer-grained lens, a higher level of stereotypes is observed toward male people when all the content related to Black people is removed from the pre-training data of the TinyBERT model (Table \ref{tbl:stereoset-subset-pre-rg}), despite the existence of many shared stereotypes between the 2 groups in stereotype benchmarks, e.g., being aggressive and athletic.
This does not occur when the same data is used to post-train the original TinyBERT model (Table \ref{tbl:stereoset-partial-post-rg}), further suggesting that side effects are not consistent across training stages.
We similarly observe unexpected side effects from models trained with DG-based stereotype mitigation.
For example, pre-training the TinyBERT model using the DG-cleaned data for the female group leads to lower stereotype levels for all the groups except for males, and pre-training on the DG-cleaned data for Black people raises the stereotype level of the model on males and Christians (Table \ref{tbl:stereoset-subset-pre-dg}).
Post-training the TinyBERT model using the DG-cleaned data for the female group, surprisingly, causes more severe stereotypes toward Christians and Muslims, 2 uncorrelated groups with the female group (Table \ref{tbl:stereoset-partial-post-dg}).

These observations reinforce our claim that the side effects of preprocessing-based stereotype mitigation can emerge across settings and are not consistently predictable from the intervention alone.

\subsection{Side Effects Persist in Massive Models} \label{sct:discussion-llama}
\begin{table}[t]
\centering
\def\arraystretch{.5}
\begin{tabular}{lccc|c}
\hline
\textbf{Category} & \textbf{LLaMa2} & \textbf{LLaMa2-RG-F} \\
\hline
Female SS       & 70.72 & \ul{66.30} \\
Male SS         & 65.48 & \ul{65.01} \\
Black SS        & 64.14 & \textbf{65.43} \\
Caucasian SS    & 67.70 & \ul{65.12} \\
Christian SS    & 69.75 & \textbf{71.03} \\
Muslim SS       & 59.50 & \ul{55.50} \\
\textbf{Overall SS}    & 64.63 & 62.97 \\
\textbf{Overall LMS}   & 90.74 & 89.71 \\
\textbf{Overall iCAT}  & 64.19 & 66.44 \\
\hline
\end{tabular}
\caption{
StereoSet results for LLaMA2 and its variant post-trained on the RG-cleaned Wikipedia corpus (female group). Underlining marks SS moving \emph{toward} 50 (less stereotyping/anti-stereotyping); bold marks SS moving \emph{away} from 50. Targeted-group improvements coincide with unintended shifts for non-targeted groups.}
\label{tbl:exp-llama}
\vspace{-15pt}
\end{table}
Since TinyBERT and GPT-2 are relatively small models, we ask whether preprocessing-based debiasing methods remain effective in larger language models, or whether the observed side effects persist at scale.
To investigate this, we conducted an additional experiment with LLaMa-2-7B (LLaMa2), a substantially larger transformer-based model than GPT-2 (124M parameters) and TinyBERT (14.5M parameters).
Given the substantial computational cost, we limited training to the RG-female setting (LLaMa2-RG-F) and evaluated the model on StereoSet (Table \ref{tbl:exp-llama}).

The results indicate that stereotypes remain prevalent in massive models.
In particular, female-oriented stereotypes were strongest in the base model (SS=70.72), motivating our focus on the female group. Post-training on RG-female data reduced the stereotype score toward females to 66.30 and produced slight reductions for the male, Caucasian, and Muslim groups. However, the same training unexpectedly increased stereotypes toward the Black and Christian groups, though these groups are not directly related to the intervention.

These findings demonstrate that the side effects of stereotype mitigation persist even in massive LLMs. This underscores the need for continued, systematic efforts to evaluate, interpret, and refine stereotype mitigation strategies to ensure that scaling up models does not simply shift or amplify biases in unpredictable ways.

\section{Conclusion \& Future Work}
\vspace{-5pt}
Stereotype mitigation is central to the safe deployment of PLMs. Preprocessing-based approaches are appealing because they remove stereotypical content before it can be learned, yet our experiments show that removing material targeting a single group from pre- or post-training data can induce unintended cross-group spillovers: stereotypes toward other, sometimes unrelated, groups may increase. These effects vary across categories, corpora, and debiasing recipes and are difficult to predict, which undermines the reliability of preprocessing as a primary mitigation strategy. Although preprocessing can improve targeted metrics, unmonitored side effects can erode reliability; data debiasing should therefore be treated as an intervention whose side effects are monitored alongside benefits. Future work includes (i) moving beyond attention-based attribution toward causal and distributional analyses of how interventions reshape co-occurrences and inductive biases, and (ii) expanding evaluation beyond StereoSet/CrowS-Pairs to additional identity axes and languages.

\section*{Ethics Statement}
We examine the side effects of data-level debiasing without releasing any post-trained models. We report only aggregate statistics to reduce risk. We curated corpora from the June 1, 2023, English Wikipedia snapshot and evaluated them on publicly released benchmarks (StereoSet; CrowS-Pairs). We used constrained LLM prompts for swapping references and removed generations containing slurs or identity-targeted insults. Because debiasing can redistribute harm, we recommend documenting both targeted and non-targeted groups in model cards and auditing side effects before deployment.

Regarding the use of generative AI (GenAI) tools, we acknowledge the use of these tools for assistance in improving the clarity and readability of the manuscript text. The use of these tools was limited to language refinement; no GenAI models were used for ideation, experimental design, or any other substantive aspects of the research.

\section*{Limitations}
We focus on the English Wikipedia and six identity groups. The effects may differ for other identities, domains, and languages. Most experiments use compact models (TinyBERT, GPT-2) for control and tractability, and we include only a single post-training slice for a larger decoder-only model. We leave broader coverage to future work. We fix the random seed to 42 for reproducibility and do not report multi-seed uncertainty for SS/CrowS-Pairs, so claims are calibrated to directional trends rather than significance tests. Finally, our DG detector, although high-recall, may include false positives, and SR swaps may introduce subtle distributional artifacts, despite manual spot checks.

We acknowledge these as important directions for future work and encourage follow-up studies to build on our findings with broader model coverage and more diverse representation.

\section*{Acknowledgments}
We are grateful to the Templeton Foundation and Georgia State University for supporting this work.

\bibliography{custom}
\cleardoublepage
\appendix
\section{Stereotype Evaluation Results} \label{appendix:stereo-results}
\renewcommand{\thefigure}{A\arabic{figure}}
\renewcommand{\thetable}{A\arabic{table}}
\setcounter{figure}{0} 
\setcounter{table}{0} 

This section details the stereotype evaluation results of all the models in our experiments.

\subsection{StereoSet} 
\label{appendix:stereo-eval:stereoset}

\subsubsection{Pre-trained Models} 
\label{appendix:stereo-eval:stereoset:pre}
Tables \ref{tbl:stereoset-full-pre-dg}–\ref{tbl:stereoset-full-pre-sr} present the StereoSet evaluation results for TinyBERT models pre-trained on the full Wikipedia corpus debiased under the DG, RG, and SR settings, respectively. Results for TinyBERT models pre-trained on a subset of the corpus (5\%) are shown in Tables \ref{tbl:stereoset-subset-pre-dg}–\ref{tbl:stereoset-subset-pre-sr}.

Similarly, the StereoSet evaluation results for GPT-2 models pre-trained on the full preprocessed corpus are reported in Tables \ref{tbl:gpt2-stereoset-full-pre-dg}–\ref{tbl:gpt2-stereoset-full-pre-sr}. Since GPT-2 training failed to converge on the 5\% subset of Wikipedia, we omit results for that setting.

\begin{table*}[t]
\centering
\def\arraystretch{.5}
\begin{tabular}{lcccccc|c}
\hline
\textbf{Stereotype Category} & \textbf{Female} & \textbf{Male} & \textbf{Black} & \textbf{Caucasian} & \textbf{Christian} & \textbf{Muslim} & \textbf{Original} \\
\hline
gender-female SS       & \ul{49.96} & \textbf{51.89} & \textbf{52.63} & \textbf{51.56} & \textbf{53.00} & \textbf{52.02} & 51.53 \\
gender-male SS         & \ul{50.59} & \ul{49.59} & \ul{49.14} & \ul{48.97} & \ul{49.67} & \ul{51.76} & 51.80 \\
race-black SS          & \textbf{53.64} & \textbf{56.51} & \ul{51.44} & \ul{52.23} & \ul{51.17} & \ul{53.07} & 53.17 \\
race-caucasian SS      & \textbf{55.28} & \textbf{53.62} & \ul{53.37} & \ul{53.07} & \textbf{55.65} & \textbf{56.05} & 53.43 \\
religion-christian SS  & \ul{57.72} & 60.19 & \ul{51.63} & \ul{56.48} & \ul{57.72} & \ul{56.48} & 60.19 \\
religion-muslim SS     & \ul{60.92} & \ul{60.79} & \ul{62.16} & \ul{62.19} & \ul{61.43} & \ul{61.14} & 64.28 \\
\textbf{Overall SS}    & 54.63 & 54.32 & 54.10 & 54.17 & 54.94 & 54.83 & 54.69 \\
\textbf{Overall LMS}  & 77.39 & 78.83 & 78.68 & 78.91 & 78.41 & 77.83 & 78.69 \\
\textbf{Overall iCAT Score} & 70.22 & 72.02 & 72.24 & 72.33 & 70.66 & 70.30 & 71.31 \\
\hline
\end{tabular}
\caption{\label{tbl:stereoset-full-pre-dg}
StereoSet evaluation results of TinyBERT models pre-trained on DG-cleaned data for each group. SS, LMS, and iCAT score indicate stereotype score, language modeling score, and idealized context association test score, respectively. Group-specific SS in the debiased models that are closer to or farther away from 50 than in the original model are underlined and bolded, respectively. Original refers to the original release of TinyBERT model without stereotype mitigation.}
\end{table*}

\begin{table*}[t]
\centering
\def\arraystretch{.5}
\begin{tabular}{lcccccc|c}
\hline
\textbf{Stereotype Category} & \textbf{Female} & \textbf{Male} & \textbf{Black} & \textbf{Caucasian} & \textbf{Christian} & \textbf{Muslim} & \textbf{Original} \\
\hline
gender-female SS       & \textbf{47.78} & \ul{50.46} & \textbf{53.70} & 51.32 & \textbf{48.55} & \textbf{52.56} & 51.53 \\
gender-male SS         & \ul{51.14} & \textbf{45.87} & \ul{48.48} & \ul{49.27} & \ul{50.71} & \textbf{52.34} & 51.80 \\
race-black SS          & \ul{49.45} & \textbf{54.29} & \ul{50.23} & \ul{51.88} & \textbf{53.63} & \ul{53.00} & 53.17 \\
race-caucasian SS      & \ul{52.82} & \ul{52.65} & \textbf{54.01} & \ul{52.99} & \textbf{55.68} & \textbf{54.85} & 53.43 \\
religion-christian SS  & \ul{56.57} & \ul{58.95} & \ul{57.67} & \ul{57.67} & \ul{48.10} & \textbf{61.42} & 60.19 \\
religion-muslim SS     & \ul{59.42} & \ul{60.56} & \ul{62.97} & \ul{61.22} & \textbf{64.34} & \ul{57.06} & 64.28 \\

\textbf{Overall SS}    & 53.78 & 53.32 & 54.16 & 54.47 & 54.88 & 54.62 & 54.69 \\
\textbf{Overall LMS}  & 76.67 & 77.04 & 78.15 & 77.39 & 77.36 & 78.24 & 78.69 \\
\textbf{Overall iCAT Score} & 70.88 & 71.93 & 71.64 & 70.48 & 69.81 & 71.01 & 71.31 \\
\hline
\end{tabular}
\caption{\label{tbl:stereoset-full-pre-rg}
StereoSet evaluation results of TinyBERT models pre-trained on RG-cleaned data for each group.}
\end{table*}

\begin{table*}[t]
\centering
\def\arraystretch{.5}
\setlength{\tabcolsep}{2pt}
\begin{tabular}{lccc|c}
\hline
\textbf{Category} & \textbf{Gender} & \textbf{Race} & \textbf{Religion} & \textbf{Original} \\
\hline
Female SS       & \ul{49.60} & \ul{50.75} & \textbf{53.67} & 51.53 \\
Male SS         & \ul{49.59} & \ul{50.71} & \textbf{51.62} & 51.80 \\
Black SS        & \textbf{54.48} & \ul{49.67} & \textbf{55.24} & 53.17 \\
Caucasian SS    & \textbf{55.84} & \textbf{52.46} & \textbf{53.53} & 53.43 \\
Christian SS    & \textbf{57.76} & \textbf{58.86} & \ul{50.53} & 60.19 \\
Muslim SS       & \textbf{64.35} & \ul{56.52} & \ul{58.15} & 64.28 \\
\textbf{Overall SS}    & 55.08 & 53.89 & 54.81 & 54.69 \\
\textbf{Overall LMS}  & 76.60 & 77.59 & 79.01 & 78.69 \\
\textbf{Overall iCAT} & 68.82 & 71.55 & 71.41 & 71.31 \\
\hline
\end{tabular}
\caption{\label{tbl:stereoset-full-pre-sr}
StereoSet evaluation results of TinyBERT models pre-trained on SR-cleaned data for each stereotype category.}
\end{table*}

\begin{table*}[t]
\centering
\def\arraystretch{.5}
\begin{tabular}{lcccccc|c}
\hline
\textbf{Stereotype Category} & \textbf{Female} & \textbf{Male} & \textbf{Black} & \textbf{Caucasian} & \textbf{Muslim} & \textbf{Christian} & \textbf{Original} \\
\hline
gender-female SS        & \ul{49.25} & \ul{48.70} & \ul{51.14} & \ul{48.38} & \ul{50.24} & \ul{48.94} & 51.73 \\
gender-male SS          & \ul{50.15} & \ul{51.76} & \ul{49.43} & \ul{51.88} & \ul{51.83} & \textbf{52.16} & 48.10 \\
race-black SS           & \ul{50.59} & \ul{50.93} & \ul{50.15} & \ul{50.83} & \ul{49.98} & \ul{46.11} & 55.31 \\
race-caucasian SS       & \ul{50.62} & \ul{52.03} & \ul{52.77} & \ul{50.11} & \ul{51.90} & \ul{50.36} & 53.84 \\
religion-muslim SS      & \ul{55.87} & \ul{56.77} & \ul{56.11} & \ul{51.46} & \ul{52.65} & \ul{53.31} & 59.78 \\
religion-christian SS   & \ul{55.38} & \ul{54.14} & \textbf{57.67} & \textbf{57.72} & \ul{55.25} & \ul{50.37} & 56.57 \\
\textbf{Overall SS}    & 52.36 & 51.94 & 52.23 & 51.76 & 52.19 & 51.65 & 52.89 \\
\textbf{Overall LMS}  & 64.60 & 64.51 & 66.24 & 64.81 & 65.17 & 62.01 & 63.14 \\
\textbf{Overall iCAT Score} & 61.55 & 61.99 & 63.23 & 62.53 & 62.33 & 59.96 & 59.50 \\
\hline
\end{tabular}
\caption{\label{tbl:stereoset-subset-pre-dg}
StereoSet evaluation results of TinyBERT models pre-trained on 5\% DG-cleaned data for each group.}
\end{table*}

\begin{table*}[t]
\centering
\def\arraystretch{.5}
\begin{tabular}{lcccccc|c}
\hline
\textbf{Stereotype Category} & \textbf{Female} & \textbf{Male} & \textbf{Black} & \textbf{Caucasian} & \textbf{Muslim} & \textbf{Christian} & \textbf{Original} \\
\hline
gender-female SS        & \textbf{41.41} & \ul{49.77} & \ul{51.00} & \textbf{48.11} & \ul{48.50} & \ul{49.74} & 51.73 \\
gender-male SS          & \textbf{53.97} & \textbf{47.25} & \ul{49.59} & \ul{50.48} & \ul{50.72} & \ul{50.89} & 48.10 \\
race-black SS           & \ul{46.63} & \ul{49.51} & \ul{47.92} & \ul{46.21} & \ul{50.78} & \ul{50.43} & 55.31 \\
race-caucasian SS       & \ul{51.51} & \ul{50.19} & \ul{53.00} & \ul{49.37} & \ul{51.63} & \ul{52.37} & 53.84 \\
religion-muslim SS      & \ul{55.43} & \ul{57.75} & \ul{54.46} & \ul{59.33} & \ul{46.83} & \ul{58.23} & 59.78 \\
religion-christian SS   & \ul{52.91} & \ul{52.91} & \textbf{58.86} & \ul{49.12} & \textbf{58.99} & \ul{45.77} & 56.57 \\
\textbf{Overall SS}    & 51.24 & 51.20 & 52.40 & 51.03 & 51.57 & 53.04 & 52.89 \\
\textbf{Overall LMS}  & 63.12 & 63.91 & 66.31 & 66.42 & 65.15 & 63.35 & 63.14 \\
\textbf{Overall iCAT Score} & 61.55 & 62.38 & 63.12 & 65.05 & 63.10 & 59.50 & 59.50 \\
\hline
\end{tabular}
\caption{\label{tbl:stereoset-subset-pre-rg}
StereoSet evaluation results of TinyBERT models pre-trained on 5\% RG-cleaned data for each group.}
\end{table*}

\begin{table*}[t]
\centering
\def\arraystretch{.5}
\setlength{\tabcolsep}{2pt}
\begin{tabular}{lccc|c}
\hline
\textbf{Category} & \textbf{Gender} & \textbf{Race} & \textbf{Religion} & \textbf{Original} \\
\hline
Female SS       & \textbf{46.55} & \ul{48.82} & \ul{50.24} & 51.73 \\
Male SS         & \ul{50.54} & \textbf{47.53} & \ul{48.75} & 48.10 \\
Black SS        & \ul{48.77} & \ul{46.80} & \ul{48.86} & 55.31 \\
Caucasian SS    & \ul{52.06} & \ul{48.67} & \ul{50.75} & 53.84 \\
Christian SS    & \ul{56.53} & \textbf{58.95} & \ul{48.10} & 56.57 \\
Muslim SS       & \ul{56.03} & \ul{55.46} & \ul{56.41} & 59.78 \\
\textbf{Overall SS}    & 51.39 & 50.96 & 51.82 & 52.89 \\
\textbf{Overall LMS}  & 64.17 & 65.60 & 63.58 & 63.14 \\
\textbf{Overall iCAT} & 62.39 & 64.34 & 61.27 & 59.50 \\
\hline
\end{tabular}
\caption{\label{tbl:stereoset-subset-pre-sr}
StereoSet evaluation results of TinyBERT models pre-trained on 5\% SR-cleaned data for each stereotype category.}
\vspace{-5pt}
\end{table*}

\begin{table*}[t]
\centering
\def\arraystretch{.5}
\begin{tabular}{lcccccc|c}
\hline
\textbf{Category Type} & \textbf{Female} & \textbf{Male} & \textbf{Black} & \textbf{Caucasian} & \textbf{Christian} & \textbf{Muslim} & \textbf{Original} \\
\hline
gender-female SS       & \ul{55.75} & \ul{55.52} & \ul{55.18} & \ul{54.99} & \textbf{56.29} & \ul{56.11} & 56.27 \\
gender-male SS         & \textbf{57.79} & \ul{54.38} & \ul{55.79} & \ul{56.31} & \textbf{57.12} & \textbf{57.64} & 56.97 \\
race-black SS          & \ul{55.08} & \ul{52.75} & \ul{53.29} & \ul{55.14} & \ul{54.55} & \ul{52.34} & 55.59 \\
race-caucasian SS      & \textbf{54.32} & \ul{53.57} & \ul{53.22} & \ul{53.55} & \ul{54.04} & \ul{52.57} & 54.14 \\
religion-christian SS  & \ul{53.09} & \ul{58.02} & \textbf{60.36} & \ul{57.94} & \ul{57.94} & \textbf{61.55} & 60.27 \\
religion-muslim SS     & \textbf{57.64} & \textbf{56.28} & \ul{54.85} & \ul{54.96} & \textbf{57.39} & \ul{55.44} & 55.99 \\
Overall SS             & 55.21 & 54.80 & 54.59 & 54.95 & 55.29 & 54.40 & 55.64 \\
Overall LMS            & 82.25 & 82.74 & 83.52 & 82.88 & 83.13 & 83.07 & 83.32 \\
Overall iCAT Score     & 73.69 & 74.79 & 75.85 & 74.68 & 74.34 & 75.75 & 73.93 \\
\hline
\end{tabular}
\caption{\label{tbl:gpt2-stereoset-full-pre-dg}StereoSet evaluation results of GPT-2 models pre-trained on full DG-cleaned Wikipedia data for each group.}
\end{table*}

\begin{table*}[t]
\centering
\def\arraystretch{.5}
\begin{tabular}{lcccccc|c}
\hline
\textbf{Category Type} & \textbf{Female} & \textbf{Male} & \textbf{Black} & \textbf{Caucasian} & \textbf{Christian} & \textbf{Muslim} & \textbf{Original} \\
\hline
gender-female SS       & \ul{49.67} & \ul{54.62} & \ul{55.61} & \textbf{56.34} & \ul{55.46} & \ul{54.74} & 56.27 \\
gender-male SS         & \ul{56.95} & \ul{52.53} & \ul{56.10} & \textbf{57.68} & \textbf{57.52} & \textbf{58.55} & 56.97 \\
race-black SS          & \ul{54.50} & \ul{54.60} & \ul{52.52} & \ul{53.41} & \ul{54.82} & \ul{55.37} & 55.59 \\
race-caucasian SS      & \textbf{54.58} & \ul{52.95} & \ul{54.09} & \ul{53.16} & \ul{53.27} & \ul{53.93} & 54.14 \\
religion-christian SS  & \ul{54.28} & \ul{53.13} & \ul{57.94} & \textbf{61.55} & \ul{54.32} & \ul{56.75} & 60.27 \\
religion-muslim SS     & \textbf{57.33} & \ul{55.17} & \ul{55.28} & \ul{54.18} & \textbf{57.83} & \ul{53.93} & 55.99 \\
Overall SS             & 54.31 & 54.36 & 55.20 & 55.02 & 55.38 & 55.24 & 55.64 \\
Overall LMS            & 80.81 & 80.25 & 83.32 & 82.94 & 82.91 & 82.91 & 83.32 \\
Overall iCAT Score     & 73.84 & 73.25 & 74.66 & 74.61 & 73.99 & 74.23 & 73.93 \\
\hline
\end{tabular}
\caption{\label{tbl:gpt2-stereoset-full-pre-rg}StereoSet evaluation results of GPT-2 models pre-trained on full RG-cleaned Wikipedia data for each group.}
\end{table*}
\begin{table*}[t]
\centering
\def\arraystretch{.5}
\begin{tabular}{lccc|c}
\hline
\textbf{Category Type} & \textbf{Gender} & \textbf{Race} & \textbf{Religion} & \textbf{Original} \\
\hline
gender-female SS       & \ul{52.91} & \ul{55.69} & \ul{54.57} & 56.27 \\
gender-male SS         & \ul{55.42} & \textbf{57.56} & \textbf{57.74} & 56.97 \\
race-black SS          & \ul{55.39} & \ul{53.33} & \ul{54.61} & 55.59 \\
race-caucasian SS      & \ul{53.91} & \ul{53.58} & \ul{53.14} & 54.14 \\
religion-christian SS  & \ul{59.13} & 60.27 & \ul{55.51} & 60.27 \\
religion-muslim SS     & \ul{55.61} & \textbf{56.37} & \textbf{59.18} & 55.99 \\
Overall SS             & 54.66 & 55.03 & 55.14 & 55.64 \\
Overall LMS            & 81.47 & 82.97 & 83.55 & 83.32 \\
Overall iCAT Score     & 73.87 & 74.63 & 74.95 & 73.93 \\
\hline
\end{tabular}
\caption{\label{tbl:gpt2-stereoset-full-pre-sr}StereoSet evaluation results of GPT-2 models pre-trained on full SR-cleaned Wikipedia data.}
\end{table*}

\subsubsection{Post-trained Models} \label{appendix:stereo-eval:stereoset:post}
Tables \ref{tbl:stereoset-full-post-dg}–\ref{tbl:stereoset-full-post-sr} report the StereoSet evaluation results for TinyBERT models post-trained on the fully preprocessed Wikipedia corpus, while Tables \ref{tbl:stereoset-partial-post-dg}–\ref{tbl:stereoset-partial-post-sr} show results for models post-trained on the 5\% subset.
For GPT-2, StereoSet evaluation results on the fully preprocessed corpus are presented in Tables \ref{tbl:gpt2-stereoset-full-post-dg}–\ref{tbl:gpt2-stereoset-full-post-sr}, and results on the 5\% subset are provided in Tables \ref{tbl:gpt2-stereoset-partial-post-dg}–\ref{tbl:gpt2-stereoset-partial-post-sr}.

\begin{table*}[t]
\centering
\def\arraystretch{.5}
\begin{tabular}{lcccccc|c}
\hline
\textbf{Stereotype Category} & \textbf{Female} & \textbf{Male} & \textbf{Black} & \textbf{Caucasian} & \textbf{Christian} & \textbf{Muslim} & \textbf{Original} \\
\hline
gender-female SS        & \ul{50.19} & \textbf{53.82} & \ul{52.13} & \ul{52.94} & \ul{51.60} & \ul{51.04} & 53.08 \\
gender-male SS          & \ul{51.34} & \ul{48.78} & \textbf{51.88} & \ul{51.20} & \textbf{52.98} & \textbf{52.41} & 51.48 \\
race-black SS           & \ul{54.45} & \ul{52.12} & \ul{52.74} & \ul{52.37} & \ul{53.67} & \ul{51.75} & 54.66 \\
race-caucasian SS       & \ul{54.13} & \ul{53.36} & \textbf{54.17} & \ul{52.96} & \ul{52.83} & \textbf{55.20} & 54.17 \\
religion-christian SS   & \textbf{58.86} & \ul{54.06} & \ul{51.68} & \ul{52.87} & \ul{54.01} & \ul{54.06} & 56.53 \\
religion-muslim SS      & \textbf{61.48} & \ul{59.89} & \ul{60.28} & \ul{59.94} & \ul{57.87} & \ul{57.30} & 61.00 \\
\textbf{Overall SS}    & 54.34 & 54.08 & 54.57 & 54.19 & 54.25 & 54.39 & 55.13 \\
\textbf{Overall LMS}  & 78.68 & 79.25 & 79.77 & 80.29 & 80.03 & 79.35 & 79.98 \\
\textbf{Overall iCAT Score} & 71.84 & 72.78 & 72.49 & 73.56 & 73.23 & 72.37 & 71.77 \\
\hline
\end{tabular}
\caption{\label{tbl:stereoset-full-post-dg}
StereoSet evaluation results of TinyBERT models post-trained on DG-cleaned data for each group.}
\end{table*}

\begin{table*}[t]
\centering
\def\arraystretch{.5}
\begin{tabular}{lcccccc|c}
\hline
\textbf{Stereotype Category} & \textbf{Female} & \textbf{Male} & \textbf{Black} & \textbf{Caucasian} & \textbf{Christian} & \textbf{Muslim} & \textbf{Original} \\
\hline
gender-female SS        & \ul{49.73} & \textbf{53.10} & \ul{50.47} & \ul{50.31} & \ul{52.87} & \ul{51.04} & 53.08 \\
gender-male SS          & \textbf{51.98} & \textbf{47.50} & \ul{50.27} & \ul{50.26} & \ul{48.58} & \textbf{52.41} & 51.48 \\
race-black SS           & \ul{54.52} & \ul{54.27} & \ul{51.13} & \ul{51.74} & \ul{54.38} & \ul{51.75} & 54.66 \\
race-caucasian SS       & \ul{53.85} & \textbf{54.21} & \ul{54.13} & \ul{52.90} & \textbf{54.54} & \textbf{55.20} & 54.17 \\
religion-christian SS   & \textbf{60.10} & \ul{51.63} & \ul{54.06} & \textbf{63.71} & \ul{46.96} & \ul{54.06} & 56.53 \\
religion-muslim SS      & \ul{60.79} & \ul{59.81} & \textbf{61.67} & \ul{60.44} & \textbf{61.60} & \ul{57.30} & 61.00 \\
\textbf{Overall SS}    & 54.38 & 53.88 & 54.38 & 53.70 & 54.65 & 54.39 & 55.13 \\
\textbf{Overall LMS}  & 78.73 & 79.14 & 80.24 & 79.73 & 79.66 & 79.35 & 79.98 \\
\textbf{Overall iCAT Score} & 71.84 & 73.01 & 73.22 & 73.83 & 72.26 & 72.37 & 71.77 \\
\hline
\end{tabular}
\caption{\label{tbl:stereoset-full-post-rg}
StereoSet evaluation results of TinyBERT models post-trained on RG-cleaned data for each group.}
\end{table*}

\begin{table*}[t]
\centering
\def\arraystretch{.5}
\setlength{\tabcolsep}{2pt}
\begin{tabular}{lccc|c}
\hline
\textbf{Category} & \textbf{Gender} & \textbf{Race} & \textbf{Religion} & \textbf{Original} \\
\hline
Female SS       & \ul{50.09} & \ul{52.70} & \ul{51.87} & 53.08 \\
Male SS         & \ul{49.13} & \textbf{45.76} & \ul{51.32} & 51.48 \\
Black SS        & \ul{53.56} & \ul{49.85} & \ul{54.13} & 54.66 \\
Caucasian SS    & \ul{53.42} & \ul{51.95} & \textbf{54.26} & 54.17 \\
Christian SS    & \ul{55.29} & \ul{53.97} & \ul{51.68} & 56.53 \\
Muslim SS       & \ul{60.99} & \textbf{62.46} & \ul{58.23} & 61.00 \\
\textbf{Overall SS}    & 53.67 & 53.71 & 54.35 & 55.13 \\
\textbf{Overall LMS}  & 79.60 & 79.99 & 80.11 & 79.98 \\
\textbf{Overall iCAT} & 73.75 & 74.06 & 73.14 & 71.77 \\
\hline
\end{tabular}
\caption{\label{tbl:stereoset-full-post-sr}
StereoSet evaluation results of TinyBERT models post-trained on SR-cleaned data.}
\vspace{-5pt}
\end{table*}

\begin{table*}[t]
\centering
\def\arraystretch{.5}
\begin{tabular}{lcccccc|c}
\hline
\textbf{Stereotype Category} & \textbf{Female} & \textbf{Male} & \textbf{Black} & \textbf{Caucasian} & \textbf{Christian} & \textbf{Muslim} & \textbf{Original} \\
\hline
gender-female SS        & \ul{52.79} & \ul{52.70} & \textbf{53.52} & \textbf{53.53} & \textbf{53.58} & \textbf{53.00} & 52.93 \\
gender-male SS          & \textbf{52.07} & \ul{49.48} & \textbf{52.42} & \ul{48.37} & \ul{51.22} & \textbf{54.58} & 51.73 \\
race-black SS           & \ul{51.95} & \ul{51.10} & \ul{52.85} & \ul{50.56} & \ul{52.74} & \ul{51.45} & 56.12 \\
race-caucasian SS       & \textbf{56.16} & \textbf{55.82} & \textbf{56.18} & \ul{54.74} & \ul{54.61} & \textbf{55.74} & 54.95 \\
religion-christian SS   & \ul{57.67} & \textbf{58.86} & \textbf{60.05} & \textbf{58.91} & \ul{56.48} & \ul{56.48} & 58.82 \\
religion-muslim SS      & \ul{60.02} & \ul{57.06} & \ul{58.66} & \textbf{60.96} & \ul{57.15} & \ul{60.07} & 60.09 \\
\textbf{Overall SS}   & 54.93 & 54.71 & 54.82 & 54.48 & 54.52 & 55.05 & 54.90 \\
\textbf{Overall LMS}  & 79.60 & 79.72 & 79.70 & 79.85 & 79.35 & 79.36 & 80.22 \\
\textbf{Overall iCAT Score} & 71.75 & 72.21 & 72.02 & 72.69 & 72.17 & 71.34 & 72.36 \\
\hline
\end{tabular}
\caption{\label{tbl:stereoset-partial-post-dg}
StereoSet evaluation results of TinyBERT models post-trained on 5\% DG-cleaned data for each group.}

\end{table*}

\begin{table*}[t]
\centering
\def\arraystretch{.5}
\begin{tabular}{lcccccc|cc}
\hline
\textbf{Stereotype Category} & \textbf{Female} & \textbf{Male} & \textbf{Black} & \textbf{Caucasian} & \textbf{Christian} & \textbf{Muslim} & \textbf{Original} \\
\hline
gender-female SS    & \ul{52.67} & \textbf{53.10} & \textbf{53.94} & \textbf{53.03} & \ul{52.20} & \textbf{54.42} & 52.93 \\
gender-male SS      & \ul{49.52} & \textbf{48.15} & \textbf{51.93} & \textbf{53.67} & \textbf{54.29} & \ul{51.49} & 51.73 \\
race-black SS       & \ul{51.94} & \ul{54.54} & \ul{52.49} & \ul{53.08} & \ul{54.23} & \ul{51.44} & 56.12 \\
race-caucasian SS   & \textbf{55.35} & \textbf{55.91} & \textbf{55.33} & \ul{54.57} & \textbf{55.84} & \textbf{55.12} & 54.95 \\
religion-christian SS & \textbf{61.24} & \ul{56.48} & \textbf{61.33} & \ul{55.29} & \ul{51.76} & \textbf{58.91} & 58.82 \\
religion-muslim SS  & \ul{57.37} & \ul{59.47} & \ul{56.54} & \ul{58.14} & \textbf{62.17} & \ul{49.78} & 60.09 \\
\textbf{Overall SS}   & 53.92 & 54.62 & 55.34 & 54.87 & 55.55 & 54.72 & 54.90 \\
\textbf{Overall LMS}  & 78.46 & 78.94 & 79.71 & 79.01 & 79.35 & 79.61 & 80.22 \\
\textbf{Overall iCAT Score} & 72.31 & 71.64 & 71.19 & 71.31 & 70.54 & 72.09 & 72.36 \\
\hline
\end{tabular}
\caption{\label{tbl:stereoset-partial-post-rg}
StereoSet evaluation results for TinyBERT models post-trained on 5\% RG-cleaned data for each group.}
\vspace{-5pt}
\end{table*}

\begin{table*}[t]
\centering
\def\arraystretch{.5}
\setlength{\tabcolsep}{2pt}
\begin{tabular}{lccc|c}
\hline
\textbf{Category} & \textbf{Gender} & \textbf{Race} & \textbf{Religion} & \textbf{Original} \\
\hline
Female SS       & \ul{52.38} & \textbf{54.04} & \ul{52.34} & 52.93 \\
Male SS         & \ul{51.17} & \textbf{53.50} & \textbf{53.86} & 51.73 \\
Black SS        & \ul{53.13} & \ul{48.72} & \ul{53.78} & 56.12 \\
Caucasian SS    & \textbf{56.24} & \ul{54.40} & \textbf{56.43} & 54.95 \\
Christian SS    & \ul{57.67} & \textbf{60.05} & \ul{55.46} & 58.82 \\
Muslim SS       & \ul{59.64} & \ul{57.59} & \ul{57.86} & 60.09 \\
\textbf{Overall SS}   & 54.77 & 54.58 & 55.65 & 54.90 \\
\textbf{Overall LMS}  & 78.81 & 79.27 & 79.49 & 80.22 \\
\textbf{Overall iCAT} & 71.29 & 71.99 & 70.50 & 72.36 \\
\hline
\end{tabular}
\caption{\label{tbl:stereoset-partial-post-sr}
StereoSet evaluation results of TinyBERT models post-trained on 5\% SR-cleaned data for each stereotype category.}
\vspace{-8pt}
\end{table*}

\begin{table*}[t]
\centering
\def\arraystretch{.5}
\begin{tabular}{lcccccc|c}
\hline
\textbf{Stereotype Category} & \textbf{Female} & \textbf{Male} & \textbf{Black} & \textbf{Caucasian} & \textbf{Christian} & \textbf{Muslim} & \textbf{Original} \\
\hline
gender-female SS       & \ul{60.86} & \textbf{61.89} & \textbf{61.36} & \textbf{61.44} & \textbf{61.36} & 61.08 & 61.08 \\
gender-male SS         & \textbf{63.23} & \ul{63.18} & \textbf{63.76} & \ul{63.17} & \textbf{63.44} & \textbf{63.49} & 63.21 \\
race-black SS          & \textbf{55.50} & \textbf{55.47} & \ul{54.94} & \ul{54.07} & \textbf{55.50} & \textbf{55.22} & 55.21 \\
race-caucasian SS      & \textbf{58.13} & \textbf{57.97} & \textbf{57.74} & \ul{57.09} & \textbf{57.94} & \textbf{57.95} & 57.37 \\
religion-christian SS  & 60.32 & 60.32 & 60.32 & \textbf{61.51} & 60.32 & 60.32 & 60.32 \\
religion-muslim SS     & 65.04 & \ul{64.39} & \ul{62.77} & \ul{62.34} & 65.04 & \ul{64.32} & 65.04 \\
Overall SS             & 59.60 & 59.86 & 59.56 & 59.23 & 59.72 & 59.62 & 59.61 \\
Overall LMS            & 90.31 & 90.37 & 90.39 & 90.52 & 90.44 & 90.34 & 90.39 \\
Overall iCAT Score     & 72.98 & 72.54 & 73.11 & 73.82 & 72.86 & 72.96 & 73.02 \\

\hline
\end{tabular}
\caption{\label{tbl:gpt2-stereoset-full-post-dg}StereoSet evaluation results of GPT-2 models post-trained on full DG-cleaned Wikipedia corpus for each group.}
\end{table*}
\begin{table*}[t]
\centering
\def\arraystretch{.5}
\begin{tabular}{lcccccc|c}
\hline
\textbf{Stereotype Category} & \textbf{Female} & \textbf{Male} & \textbf{Black} & \textbf{Caucasian} & \textbf{Christian} & \textbf{Muslim} & \textbf{Original} \\
\hline
gender-female SS       & \ul{60.61} & \textbf{61.88} & \textbf{61.21} & \textbf{61.21} & \textbf{61.15} & \ul{60.79} & 61.08 \\
gender-male SS         & \textbf{65.36} & \ul{62.91} & \ul{62.90} & \ul{63.17} & \ul{62.89} & \textbf{63.71} & 63.21 \\
race-black SS          & 55.21 & \textbf{55.47} & \ul{53.56} & \ul{53.29} & \ul{54.89} & \textbf{55.50} & 55.21 \\
race-caucasian SS      & \textbf{58.03} & \textbf{57.98} & \ul{57.00} & \ul{56.92} & \textbf{57.55} & \textbf{57.77} & 57.37 \\
religion-christian SS  & 60.32 & 60.32 & \textbf{61.51} & \textbf{61.46} & \ul{60.27} & 60.32 & 60.32 \\
religion-muslim SS     & \ul{63.68} & \ul{63.75} & \ul{61.44} & \ul{61.44} & 65.04 & \ul{64.32} & 65.04 \\
Overall SS             & 59.91 & 59.85 & 59.10 & 59.08 & 59.64 & 59.60 & 59.61 \\
Overall LMS            & 90.32 & 90.34 & 90.55 & 90.56 & 90.55 & 90.41 & 90.39 \\
Overall iCAT Score     & 72.41 & 72.54 & 74.08 & 74.10 & 73.10 & 73.04 & 73.02 \\

\hline
\end{tabular}
\caption{\label{tbl:gpt2-stereoset-full-post-rg}StereoSet evaluation results of GPT-2 models post-trained on full RG-cleaned Wikipedia corpus for each group.}
\end{table*}
\begin{table*}[t]
\centering
\def\arraystretch{.5}
\begin{tabular}{lccc|c}
\hline
\textbf{Stereotype Category} & \textbf{Gender} & \textbf{Race} & \textbf{Religion} & \textbf{Original} \\
\hline
gender-female SS       & \ul{58.26} & \ul{60.64} & \textbf{61.36} & 61.08 \\
gender-male SS         & \ul{62.66} & \textbf{63.44} & \ul{62.91} & 63.21 \\
race-black SS          & \ul{53.59} & \ul{54.42} & \textbf{55.48} & 55.21 \\
race-caucasian SS      & \textbf{57.58} & \ul{56.70} & \textbf{58.11} & 57.37 \\
religion-christian SS  & \textbf{62.70} & \textbf{63.89} & \textbf{61.51} & 60.32 \\
religion-muslim SS     & \ul{60.72} & \ul{63.10} & \ul{61.48} & 65.04 \\
Overall SS             & 58.84 & 59.29 & 59.46 & 59.61 \\
Overall LMS            & 89.97 & 90.45 & 90.30 & 90.39 \\
Overall iCAT Score     & 74.07 & 73.65 & 73.22 & 73.02 \\

\hline
\end{tabular}
\caption{\label{tbl:gpt2-stereoset-full-post-sr}StereoSet evaluation results of GPT-2 models post-trained on full SR-cleaned Wikipedia corpus.}
\end{table*}

\begin{table*}[htbp]
\centering
\begin{tabular}{lcccccc|c}
\hline
\textbf{Stereotype Category} & \textbf{Female} & \textbf{Male} & \textbf{Black} & \textbf{Caucasian} & \textbf{Christian} & \textbf{Muslim} & \textbf{Original} \\
\hline
gender-female SS        & \ul{61.56} & \textbf{62.37} & \ul{62.05} & \textbf{62.56} & \ul{61.31} & \textbf{62.77} & 62.09 \\
gender-male SS          & \textbf{62.78} & \ul{61.99} & \ul{60.17} & \textbf{64.81} & \ul{61.93} & \textbf{64.02} & 62.41 \\
race-black SS           & \textbf{54.97} & \textbf{56.74} & \ul{54.16} & \ul{53.35} & \textbf{56.26} & \textbf{55.61} & 54.94 \\
race-caucasian SS       & \ul{56.34} & \ul{56.45} & \ul{56.91} & \ul{56.52} & \textbf{57.58} & \textbf{57.40} & 57.11 \\
religion-christian SS   & \textbf{65.17} & \ul{61.51} & \textbf{66.31} & \textbf{65.12} & \ul{63.89} & \ul{63.93} & 65.08 \\
religion-muslim SS      & \ul{61.83} & \ul{62.46} & \textbf{63.85} & \textbf{66.12} & \ul{60.99} & \ul{61.92} & 62.68 \\
\textbf{Overall SS}    & 59.49 & 59.84 & 59.40 & 59.73 & 60.01 & 60.13 & 59.58 \\
\textbf{Overall LMS}  & 89.91 & 89.78 & 89.92 & 89.91 & 89.70 & 89.87 & 89.69 \\
\textbf{Overall iCAT Score} & 72.84 & 72.11 & 73.01 & 72.42 & 71.73 & 71.65 & 72.50 \\
\hline
\end{tabular}
\caption{\label{tbl:gpt2-stereoset-partial-post-dg} 
StereoSet evaluation results of GPT-2 models post-trained on 5\% DG-cleaned Wikipedia data for each group.}
\end{table*}

\begin{table*}[htbp]
\centering
\begin{tabular}{lcccccc|c}
\hline
\textbf{Stereotype Category} & \textbf{Female} & \textbf{Male} & \textbf{Black} & \textbf{Caucasian} & \textbf{Christian} & \textbf{Muslim} & \textbf{Original} \\
\hline
gender-female SS        & \ul{61.82} & \textbf{62.82} & \ul{61.58} & \textbf{62.80} & \textbf{62.34} & \textbf{62.83} & 62.09 \\
gender-male SS          & \textbf{64.07} & \ul{61.87} & \textbf{63.60} & \textbf{63.25} & \textbf{63.26} & \textbf{63.10} & 62.41 \\
race-black SS           & \textbf{56.74} & \textbf{56.96} & \ul{54.71} & \textbf{55.23} & \textbf{55.85} & \textbf{55.76} & 54.94 \\
race-caucasian SS       & \ul{56.74} & \textbf{58.00} & \ul{56.33} & \ul{56.54} & \textbf{57.62} & \textbf{58.15} & 57.11 \\
religion-christian SS   & \ul{63.89} & \ul{62.70} & \ul{63.93} & \ul{62.65} & \ul{63.93} & \ul{62.74} & 65.08 \\
religion-muslim SS      & \textbf{63.75} & \ul{62.66} & \textbf{66.19} & \textbf{66.09} & \ul{62.39} & \ul{61.34} & 62.68 \\
\textbf{Overall SS}    & 60.18 & 60.01 & 59.81 & 60.15 & 59.53 & 60.33 & 59.58 \\
\textbf{Overall LMS}  & 89.41 & 89.72 & 90.10 & 89.86 & 89.75 & 89.43 & 89.69 \\
\textbf{Overall iCAT Score} & 71.20 & 71.76 & 72.42 & 71.61 & 72.64 & 70.96 & 72.50 \\
\hline
\end{tabular}
\caption{\label{tbl:gpt2-stereoset-post-rg} 
StereoSet evaluation results of GPT-2 models post-trained on 5\% RG-cleaned Wikipedia data for each group.}
\end{table*}

\begin{table*}[htbp]
\centering
\begin{tabular}{lccc|c}
\hline
\textbf{Category} & \textbf{Gender} & \textbf{Race} & \textbf{Religion} & \textbf{Original} \\
\hline
Female SS       & \ul{60.01} & \textbf{62.87} & \ul{62.03} & 62.09 \\
Male SS         & \ul{60.09} & \textbf{64.32} & \textbf{62.62} & 62.41 \\
Black SS        & \ul{51.90} & \ul{54.14} & \ul{54.06} & 54.94 \\
Caucasian SS    & \ul{56.10} & \ul{56.33} & \ul{56.06} & 57.11 \\
Christian SS    & \textbf{66.31} & \ul{61.51} & \ul{62.74} & 65.08 \\
Muslim SS       & \ul{60.54} & \textbf{63.33} & \ul{62.09} & 62.68 \\
\textbf{Overall SS}    & 58.61 & 59.42 & 59.51 & 59.58 \\
\textbf{Overall LMS}  & 88.17 & 90.09 & 89.87 & 89.69 \\
\textbf{Overall iCAT Score} & 72.99 & 73.11 & 72.77 & 72.50 \\
\hline
\end{tabular}
\caption{\label{tbl:gpt2-stereoset-partial-post-sr} 
StereoSet evaluation results of GPT-2 models post-trained on 5\% SR-cleaned Wikipedia data for each stereotype category.}
\end{table*}

\subsection{CrowS-Pairs} \label{appendix:stereo-eval:crows}

\subsubsection{Pre-trained Models} \label{appendix:stereo-eval:crows:pre}
We show the CrowS-Pairs evaluation results of TinyBERT models pre-trained on the full preprocessed Wikipedia corpus (Tables \ref{tbl:tinybert-dg-pre-full-crows} - \ref{tbl:tinybert-sr-pre-full-crows}) and on 5\% preprocessed data (Tables \ref{tbl:tinybert-pre-partial-dg-crows} - \ref{tbl:tinybert-pre-partial-sr-crows})
Likewise, Tables \ref{tbl:gpt2-crows-full-pre-dg}–\ref{tbl:gpt2-crows-full-pre-sr} present the CrowS-Pairs evaluation results for GPT-2 models pre-trained on the fully preprocessed Wikipedia corpus.
Since GPT-2 pre-training did not converge on the 5\% subset, we omit CrowS-Pairs evaluation for those models.

\begin{table*}[t]
\centering
\def\arraystretch{.8}
\begin{tabular}{lcccccc|c}
\hline
\textbf{Category Type} & \textbf{Female} & \textbf{Male} & \textbf{Black} & \textbf{Caucasian} & \textbf{Christian} & \textbf{Muslim} & \textbf{Original} \\
\hline
gender    & 51.53 & 53.82 & 53.44 & 51.53 & 55.34 & 52.29 & 51.15 \\
race      & 61.55 & 61.75 & 59.03 & 58.83 & 60.78 & 61.75 & 65.24 \\
religion  & 60.95 & 60.00 & 61.90 & 62.86 & 62.86 & 62.86 & 46.67 \\
overall score  & 58.01 & 58.52 & 58.12 & 57.74 & 59.66 & 58.97 & 54.35 \\
\hline
\end{tabular}
\caption{\label{tbl:tinybert-dg-pre-full-crows}CrowS-Pairs evaluation results of TinyBERT models pre-trained on the full DG-cleaned Wikipedia data.}
\end{table*}

\begin{table*}[t]
\centering
\def\arraystretch{.8}
\begin{tabular}{lcccccc|c}
\hline
\textbf{Category Type} & \textbf{Female} & \textbf{Male} & \textbf{Black} & \textbf{Caucasian} & \textbf{Christian} & \textbf{Muslim} & \textbf{Original} \\
\hline
gender    & 54.58 & 51.53 & 54.58 & 48.85 & 50.76 & 54.58 & 51.15 \\
race      & 51.07 & 60.19 & 57.09 & 64.27 & 62.91 & 59.03 & 65.24 \\
religion  & 69.52 & 60.95 & 74.29 & 66.67 & 63.81 & 47.62 & 46.67 \\
overall score  & 58.39 & 57.56 & 61.99 & 59.93 & 59.16 & 53.74 & 54.35 \\
\hline
\end{tabular}
\caption{\label{tbl:tinybert-rg-pre-full-crows}CrowS-Pairs evaluation results of TinyBERT models pre-trained on the full RG-cleaned Wikipedia data.}
\end{table*}

\begin{table*}[t]
\centering
\def\arraystretch{.8}
\begin{tabular}{lccc|c}
\hline
\textbf{Category Type} & \textbf{Gender} & \textbf{Race} & \textbf{Religion} & \textbf{Original} \\
\hline
gender    & 44.27 & 52.29 & 51.53 & 51.15 \\
race      & 57.09 & 49.13 & 51.46 & 65.24 \\
religion  & 63.81 & 55.24 & 62.86 & 46.67 \\
overall score  & 55.06 & 52.22 & 55.28 & 54.35 \\
\hline
\end{tabular}
\caption{\label{tbl:tinybert-sr-pre-full-crows}CrowS-Pairs evaluation results of TinyBERT models pre-trained on the full SR-cleaned Wikipedia data.}
\end{table*}

\begin{table*}[t]
\centering
\def\arraystretch{.8}
\begin{tabular}{lcccccc|c}
\hline
\textbf{Category Type} & \textbf{Female} & \textbf{Male} & \textbf{Black} & \textbf{Caucasian} & \textbf{Christian} & \textbf{Muslim} & \textbf{Original} \\
\hline
gender    & 54.58 & 51.15 & 44.27 & 49.62 & 58.02 & 51.53 & 51.15 \\
race      & 58.06 & 62.33 & 67.77 & 61.75 & 62.33 & 58.64 & 65.24 \\
religion  & 60.95 & 53.33 & 60.00 & 60.00 & 57.14 & 56.19 & 46.67 \\
overall score  & 57.86 & 55.60 & 57.35 & 57.12 & 59.16 & 55.45 & 54.35 \\
\hline
\end{tabular}
\caption{\label{tbl:tinybert-pre-partial-dg-crows}CrowS-Pairs evaluation results of TinyBERT models pre-trained on the 5\% DG-cleaned Wikipedia data.}
\end{table*}

\begin{table*}[t]
\centering
\def\arraystretch{.8}
\begin{tabular}{lcccccc|c}
\hline
\textbf{Category Type} & \textbf{Female} & \textbf{Male} & \textbf{Black} & \textbf{Caucasian} & \textbf{Christian} & \textbf{Muslim} & \textbf{Original} \\
\hline
gender    & 53.05 & 51.91 & 53.05 & 47.71 & 53.05 & 52.67 & 51.15 \\
race      & 60.78 & 59.42 & 64.85 & 64.47 & 64.66 & 62.52 & 65.24 \\
religion  & 55.24 & 50.48 & 71.43 & 54.29 & 45.71 & 38.10 & 46.67 \\
overall score  & 56.36 & 53.94 & 63.11 & 55.49 & 54.47 & 51.10 & 54.35 \\
\hline
\end{tabular}
\caption{\label{tbl:tinybert-pre-partial-rg-crows}CrowS-Pairs evaluation results of TinyBERT models pre-trained on the 5\% RG-cleaned Wikipedia data.}
\end{table*}

\begin{table*}[t]
\centering
\def\arraystretch{.8}
\begin{tabular}{lccc|c}
\hline
\textbf{Category Type} & \textbf{Gender} & \textbf{Race} & \textbf{Religion} & \textbf{Original} \\
\hline
gender    & 46.18 & 51.15 & 51.91 & 51.15 \\
race      & 61.94 & 48.35 & 57.86 & 65.24 \\
religion  & 51.43 & 56.19 & 69.52 & 46.67 \\
overall score  & 53.18 & 51.90 & 59.76 & 54.35 \\
\hline
\end{tabular}
\caption{\label{tbl:tinybert-pre-partial-sr-crows}CrowS-Pairs evaluation results of TinyBERT models pre-trained on the 5\% SR-cleaned Wikipedia data.}
\end{table*}

\begin{table*}[t]
\centering
\def\arraystretch{.5}
\begin{tabular}{lcccccc|c}
\hline
\textbf{Category Type} & \textbf{Female} & \textbf{Male} & \textbf{Black} & \textbf{Caucasian} & \textbf{Christian} & \textbf{Muslim} & \textbf{Original} \\
\hline
gender & 48.09 & 53.05 & 53.05 & 51.53 & 52.29 & 52.29 & 51.15 \\
race & 57.36 & 59.88 & 56.78 & 58.33 & 56.2 & 56.98 & 58.53 \\
religion & 44.76 & 43.81 & 45.71 & 41.9 & 45.71 & 38.1 & 43.81 \\
overall score & 53.65 & 55.11 & 54.44 & 54.58 & 54.05 & 53.65 & 54.31 \\
\hline
\end{tabular}
\caption{\label{tbl:gpt2-crows-full-pre-dg}CrowS-Pairs evaluation results of GPT-2 models pre-trained on the full DG-cleaned Wikipedia data.}
\end{table*}

\begin{table*}[t]
\centering
\def\arraystretch{.5}
\begin{tabular}{lcccccc|c}
\hline
\textbf{Category Type} & \textbf{Female} & \textbf{Male} & \textbf{Black} & \textbf{Caucasian} & \textbf{Christian} & \textbf{Muslim} & \textbf{Original} \\
\hline
gender & 45.04 & 46.56 & 51.53 & 52.29 & 52.29 & 51.15 & 51.15 \\
race & 57.75 & 61.24 & 56.2 & 57.75 & 59.5 & 56.78 & 58.53 \\
religion & 43.81 & 39.05 & 44.76 & 43.81 & 68.57 & 36.19 & 43.81 \\
overall score & 52.79 & 52.85 & 53.65 & 54.64 & 56.96 & 53.38 & 54.31 \\
\hline
\end{tabular}
\caption{\label{tbl:gpt2-crows-full-pre-rg}CrowS-Pairs evaluation results of GPT-2 models pre-trained on the full RG-cleaned Wikipedia data.}
\end{table*}

\begin{table*}[t]
\centering
\def\arraystretch{.5}
\begin{tabular}{lccc|c}
\hline
\textbf{Category Type} & \textbf{Gender} & \textbf{Race} & \textbf{Religion} & \textbf{Original} \\
\hline
gender & 40.46 & 53.05 & 51.53 & 51.15 \\
race & 57.17 & 60.66 & 58.91 & 58.53 \\
religion & 47.62 & 47.62 & 53.33 & 43.81 \\
overall score & 51.53 & 55.77 & 55.44 & 54.31 \\
\hline
\end{tabular}
\caption{\label{tbl:gpt2-crows-full-pre-sr}CrowS-Pairs evaluation results of GPT-2 models pre-trained on the full SR-cleaned Wikipedia data.}
\end{table*}

\subsubsection{Post-trained Models} \label{appendix:stereo-eval:crows:post}
Tables \ref{tbl:tinybert-post-full-dg-crows-post}–\ref{tbl:tinybert-post-full-sr-crows-post} present the CrowS-Pairs stereotype evaluation results for TinyBERT models post-trained on the full preprocessed Wikipedia corpus, while Tables \ref{tbl:tinybert-post-partial-dg-crows-post}–\ref{tbl:tinybert-post-partial-sr-crows-post} show the corresponding results for TinyBERT models trained on a 5\% subset of the same corpus. For GPT-2 models, the evaluation results following post-training on the full preprocessed Wikipedia corpus are reported in Tables \ref{tbl:gpt2-crows-full-post-dg}–\ref{tbl:gpt2-crows-full-post-sr}, and those for models post-trained on the subsampled data are displayed in Tables \ref{tbl:gpt2-crows-partial-post-dg} - \ref{tbl:gpt2-crows-partial-post-sr}.

\begin{table*}[t]
\centering
\def\arraystretch{.8}
\begin{tabular}{lcccccc|c}
\hline
\textbf{Category Type} & \textbf{Female} & \textbf{Male} & \textbf{Black} & \textbf{Caucasian} & \textbf{Christian} & \textbf{Muslim} & \textbf{Original} \\
\hline
gender    & 51.15 & 54.96 & 50.38 & 45.04 & 53.82 & 51.53 & 51.15 \\
race      & 52.43 & 51.46 & 52.23 & 58.83 & 54.76 & 57.67 & 65.24 \\
religion  & 64.76 & 64.76 & 64.76 & 63.81 & 60.95 & 57.14 & 46.67 \\
overall score & 56.11    & 57.06    & 55.79 & 55.89 & 56.51  & 55.45   & 54.35 \\
\hline
\end{tabular}
\caption{\label{tbl:tinybert-post-full-dg-crows-post}CrowS-Pairs evaluation results of TinyBERT models post-trained on the full DG-cleaned Wikipedia data.}
\end{table*}

\begin{table*}[t]
\centering
\def\arraystretch{.8}
\begin{tabular}{lcccccc|c}
\hline
\textbf{Category Type} & \textbf{Female} & \textbf{Male} & \textbf{Black} & \textbf{Caucasian} & \textbf{Christian} & \textbf{Muslim} & \textbf{Original} \\
\hline
gender    & 53.82 & 53.44 & 50.38 & 45.42 & 52.29 & 54.58 & 51.15 \\
race      & 55.15 & 61.55 & 56.50 & 63.69 & 59.22 & 61.36 & 65.24 \\
religion  & 64.76 & 62.86 & 76.19 & 60.95 & 59.05 & 50.48 & 46.67 \\
overall score  & 57.91 & 59.28 & 61.02 & 56.69 & 56.85 & 55.47 & 54.35 \\
\hline
\end{tabular}
\caption{\label{tbl:tinybert-post-full-rg-crows-post}CrowS-Pairs evaluation results of TinyBERT models post-trained on the full RG-cleaned Wikipedia data.}
\end{table*}

\begin{table*}[t]
\centering
\def\arraystretch{.8}
\begin{tabular}{lccc|c}
\hline
\textbf{Category Type} & \textbf{Gender} & \textbf{Race} & \textbf{Religion} & \textbf{Original} \\
\hline
gender    & 43.89 & 50.00 & 52.67 & 51.15 \\
race      & 50.29 & 51.46 & 55.15 & 65.24 \\
religion  & 59.05 & 64.76 & 63.81 & 46.67 \\
overall score  & 51.08 & 55.41 & 57.21 & 54.35 \\
\hline
\end{tabular}
\caption{\label{tbl:tinybert-post-full-sr-crows-post}CrowS-Pairs evaluation results of TinyBERT models post-trained on the full SR-cleaned Wikipedia data.}
\end{table*}

\begin{table*}[t]
\centering
\def\arraystretch{.8}
\begin{tabular}{lcccccc|c}
\hline
\textbf{Category Type} & \textbf{Female} & \textbf{Male} & \textbf{Black} & \textbf{Caucasian} & \textbf{Christian} & \textbf{Muslim} & \textbf{Original} \\
\hline
gender    & 55.34 & 55.34 & 53.82 & 53.82 & 53.44 & 53.44 & 51.15 \\
race      & 56.31 & 50.29 & 58.64 & 53.40 & 54.76 & 55.15 & 65.24 \\
religion  & 63.81 & 65.71 & 68.57 & 63.81 & 62.86 & 59.05 & 46.67 \\
overall score  & 58.49 & 57.11 & 60.34 & 57.01 & 57.02 & 55.88 & 54.35 \\
\hline
\end{tabular}
\caption{\label{tbl:tinybert-post-partial-dg-crows-post}CrowS-Pairs evaluation results of TinyBERT models post-trained on the 5\% DG-cleaned Wikipedia data.}
\end{table*}

\begin{table*}[t]
\centering
\def\arraystretch{.8}
\begin{tabular}{lcccccc|c}
\hline
\textbf{Category Type} & \textbf{Female} & \textbf{Male} & \textbf{Black} & \textbf{Caucasian} & \textbf{Christian} & \textbf{Muslim} & \textbf{Original} \\
\hline
gender    & 58.02 & 54.58 & 56.49 & 46.95 & 51.91 & 55.34 & 51.15 \\
race      & 58.83 & 58.64 & 54.76 & 56.50 & 57.67 & 55.53 & 65.24 \\
religion  & 64.76 & 68.57 & 74.29 & 69.52 & 65.71 & 51.43 & 46.67 \\
overall score  & 60.54 & 60.60 & 61.85 & 57.66 & 58.43 & 54.10 & 54.35 \\
\hline
\end{tabular}
\caption{\label{tbl:tinybert-post-partial-rg-crows-post}CrowS-Pairs evaluation results of TinyBERT models post-trained on the 5\% RG-cleaned Wikipedia data.}
\end{table*}

\begin{table*}[t]
\centering
\def\arraystretch{.8}
\begin{tabular}{lccc|c}
\hline
\textbf{Category Type} & \textbf{Gender} & \textbf{Race} & \textbf{Religion} & \textbf{Original} \\
\hline
gender    & 42.75 & 55.73 & 56.11 & 51.15 \\
race      & 58.64 & 45.44 & 54.95 & 65.24 \\
religion  & 62.86 & 60.95 & 66.67 & 46.67 \\
overall score  & 54.75 & 54.04 & 59.24 & 54.35 \\
\hline
\end{tabular}
\caption{\label{tbl:tinybert-post-partial-sr-crows-post}CrowS-Pairs evaluation results of TinyBERT models post-trained on the 5\% SR-cleaned Wikipedia data.}
\end{table*}

\begin{table*}[t]
\centering
\def\arraystretch{.5}
\begin{tabular}{lcccccc|c}
\hline
\textbf{Category Type} & \textbf{Female} & \textbf{Male} & \textbf{Black} & \textbf{Caucasian} & \textbf{Christian} & \textbf{Muslim} & \textbf{Original} \\
\hline
gender & 58.78 & 53.44 & 58.78 & 59.92 & 58.78 & 59.54 & 59.16 \\
race & 58.14 & 57.75 & 58.14 & 58.53 & 58.14 & 57.95 & 58.14 \\
religion & 57.14 & 57.14 & 57.14 & 56.19 & 57.14 & 57.14 & 57.14 \\
overall score & 58.69 & 57.29 & 58.75 & 58.82 & 58.69 & 58.69 & 58.69 \\
\hline
\end{tabular}
\caption{\label{tbl:gpt2-crows-full-post-dg}CrowS-Pairs evaluation results of GPT-2 models post-trained on the full DG-cleaned Wikipedia data.}
\end{table*}

\begin{table*}[t]
\centering
\def\arraystretch{.5}
\begin{tabular}{lcccccc|c}
\hline
\textbf{Category Type} & \textbf{Female} & \textbf{Male} & \textbf{Black} & \textbf{Caucasian} & \textbf{Christian} & \textbf{Muslim} & \textbf{Original} \\
\hline
gender & 57.63 & 53.44 & 59.92 & 59.92 & 59.54 & 59.16 & 59.16 \\
race & 57.95 & 58.14 & 58.53 & 58.33 & 59.88 & 58.14 & 58.14 \\
religion & 57.14 & 57.14 & 56.19 & 56.19 & 56.19 & 56.19 & 57.14 \\
overall score & 58.09 & 57.49 & 58.82 & 58.69 & 59.22 & 58.75 & 58.69 \\
\hline
\end{tabular}
\caption{\label{tbl:gpt2-crows-full-post-rg}CrowS-Pairs evaluation results of GPT-2 models post-trained on the full RG-cleaned Wikipedia data.}
\end{table*}

\begin{table*}[t]
\centering
\def\arraystretch{.5}
\begin{tabular}{lccc|c}
\hline
\textbf{Category Type} & \textbf{Gender} & \textbf{Race} & \textbf{Religion} & \textbf{Original} \\
\hline
gender & 55.73 & 57.63 & 59.16 & 59.16 \\
race & 59.88 & 59.3 & 58.14 & 58.14 \\
religion & 55.24 & 57.14 & 57.14 & 57.14 \\
overall score & 58.16 & 58.82 & 58.75 & 58.69 \\
\hline
\end{tabular}
\caption{\label{tbl:gpt2-crows-full-post-sr}CrowS-Pairs evaluation results of GPT-2 models post-trained on the full SR-cleaned Wikipedia data.}
\end{table*}

\begin{table*}[t]
\centering
\def\arraystretch{.5}
\begin{tabular}{lcccccc|c}
\hline
\textbf{Category Type} & \textbf{Female} & \textbf{Male} & \textbf{Black} & \textbf{Caucasian} & \textbf{Christian} & \textbf{Muslim} & \textbf{Original} \\
\hline
gender & 54.20 & 57.63 & 56.87 & 54.58 & 58.02 & 53.82 & 55.34 \\
race & 59.50 & 67.05 & 61.05 & 63.57 & 63.37 & 62.60 & 62.02 \\
religion & 54.29 & 55.24 & 56.19 & 55.24 & 59.05 & 53.33 & 56.19 \\
overall score & 58.09 & 61.87 & 59.02 & 59.95 & 60.48 & 59.28 & 59.28 \\
\hline
\end{tabular}
\caption{\label{tbl:gpt2-crows-partial-post-dg}CrowS-Pairs evaluation results of GPT-2 models post-trained on the 5\% DG-cleaned Wikipedia data.}
\end{table*}

\begin{table*}[t]
\centering
\def\arraystretch{.5}
\begin{tabular}{lcccccc|c}
\hline
\textbf{Category Type} & \textbf{Female} & \textbf{Male} & \textbf{Black} & \textbf{Caucasian} & \textbf{Christian} & \textbf{Muslim} & \textbf{Original} \\
\hline
gender & 45.80 & 57.63 & 52.67 & 53.05 & 56.49 & 53.05 & 55.34 \\
race & 60.27 & 65.89 & 60.66 & 62.21 & 62.79 & 61.05 & 62.02 \\
religion & 56.19 & 56.19 & 57.14 & 55.24 & 59.05 & 54.29 & 56.19 \\
overall score & 57.56 & 61.41 & 58.16 & 59.02 & 60.08 & 58.62 & 59.28 \\
\hline
\end{tabular}
\caption{\label{tbl:gpt2-crows-partial-post-rg}CrowS-Pairs evaluation results of GPT-2 models post-trained on the 5\% RG-cleaned Wikipedia data.}
\end{table*}

\begin{table*}[t]
\centering
\def\arraystretch{.5}
\begin{tabular}{lcccc|c}
\hline
\textbf{Category Type} & \textbf{Gender} & \textbf{Race} & \textbf{Religion} & \textbf{Original} \\
\hline
gender & 51.15 & 51.91 & 53.44 & 55.34 \\
race & 62.60 & 63.57 & 61.24 & 62.02 \\
religion & 53.33 & 56.19 & 55.24 & 56.19 \\
overall score & 58.22 & 59.08 & 58.36 & 59.28 \\
\hline
\end{tabular}
\caption{\label{tbl:gpt2-crows-partial-post-sr}CrowS-Pairs evaluation results of GPT-2 models post-trained on the 5\% SR-cleaned Wikipedia data.}
\end{table*}

\end{document}